\documentclass[10pt,twocolumn,letterpaper]{article}
\usepackage[pagenumbers]{cvpr}
\usepackage{cuted}
\definecolor{cvprblue}{rgb}{0.21,0.49,0.74}
\usepackage[pagebackref,breaklinks,colorlinks,allcolors=cvprblue]{hyperref}

\def\confName{CVPR}
\def\confYear{2027}

\title{
Focus4Dex: Action-Guided Visual Information Selection and Utilization for Dexterous Manipulation
}

\author{
Wensi Huang$^{1,2,3}$ \quad
Yangkun Zhu$^{3}$ \quad
Tutian Tang$^{3,4}$ \quad
Wanli Xing$^{2}$ \quad
Xingyu Ji$^{2,4}$ \\
Yang Li$^{3}$ \quad
Feng Zhao$^{1\dagger}$ \quad
Kaifeng Zhang$^{2\dagger}$ \quad
Qiaojun Yu$^{3,4\dagger}$ \quad
Weinan Zhang$^{3,4}$ \\[4pt]
{\small $^{1}$University of Science and Technology of China \quad $^{2}$Sharpa} \\
{\small $^{3}$Shanghai Artificial Intelligence Laboratory \quad $^{4}$Shanghai Jiao Tong University} \\[3pt]
{\small $^{\dagger}$Corresponding Authors}
}

\begin{document}
\maketitle
\begin{strip}
\centering
\includegraphics[width=\textwidth]{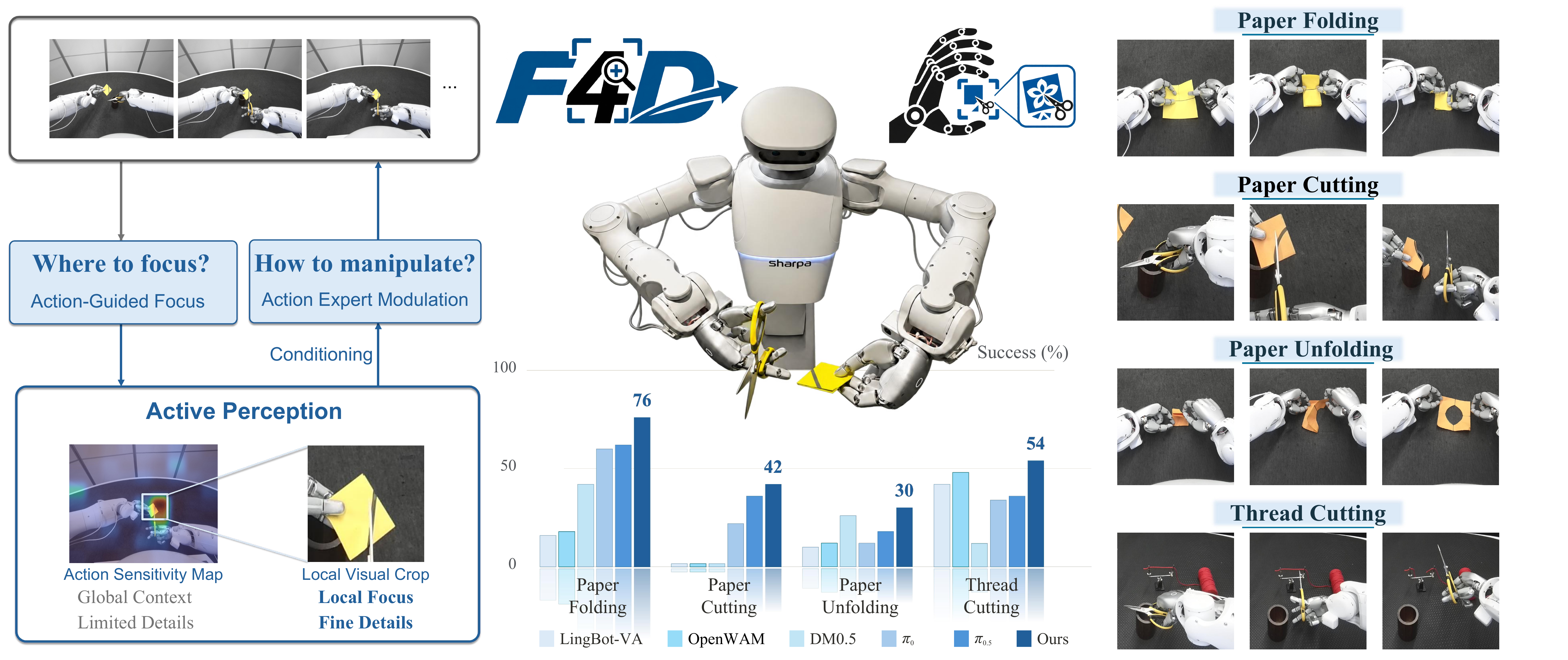}
\captionsetup{hypcap=false}
\captionof{figure}{\textbf{Action-Guided Visual Focus for Dexterous Manipulation.} (\emph{Left}) For fine-grained dexterous manipulation, Focus4Dex establishes an active perception loop addressing \emph{where to focus} and \emph{how to manipulate}. Guided by action sensitivity, the policy localizes Action-sensitive Visual Regions (AVRs) to extract high-resolution local visual crops and conditions the action expert on these detailed visual cues while preserving global scene context. (\emph{Middle \& Right}) Focus4Dex outperforms all baselines across four real-world dexterous tool-use and deformable-object manipulation tasks spanning paper folding, paper cutting, paper unfolding, and thread cutting.}
\label{figs:teaser}
\captionsetup{hypcap=true}
\end{strip}
\begin{abstract}
Recent progress in robotic hardware and foundation models has enabled sophisticated dexterous skills, including tool use and deformable object manipulation. However, while existing methods predominantly investigate tactile and force feedback in such contact-rich tasks, fine-grained visual perception remains largely underexplored. In this paper, we present Focus4Dex, bridging task-critical visual discovery with local action conditioning to resolve both \emph{where to focus} and \emph{how to manipulate} in precise dexterous manipulation. Our pipeline comprises three components: (1) Action-sensitive Visual Region (AVR) Discovery, which identifies teacher AVRs offline via localized visual perturbations; (2) an online AVR Localization Module supervised by these teacher AVRs to predict AVRs in real time without online perturbations; and (3) Action-conditioned Visual Modulation, which combines local visual features and region geometry to modulate the action expert via gated adaptive RMS normalization (AdaRMS) while preserving global scene representations. Evaluated across four real-world tasks spanning tool use and deformable object manipulation, Focus4Dex achieves a 50.5\% mean success rate, outperforming the strongest baseline by 12.5 percentage points. Comprehensive ablations further verify the effectiveness of our conditioning design. Project website: \url{https://0309hws.github.io/Focus4Dex.github.io/}
\end{abstract}

\section{Introduction}
\label{sec:introduction}
Featuring a human-inspired design, dexterous robotic hands are well suited for everyday tasks requiring skillful tool use and precise manipulation of deformable objects~\cite{dexcap2024,grasp2dexterity2026,origami2021,dexumi2025,dexknot2026}. Advancing these capabilities is an important step toward deploying robots in unstructured human-centric environments such as homes and workplaces. In these settings, robots should reliably operate tools designed for humans and handle everyday materials to provide practical assistance~\cite{pi05}. Nevertheless, mastering such fine-grained skills demands both accurate perception of spatial relationships and precise regulation of tool--object interactions, where even minor misalignments or contact \mbox{errors can lead to task failure~\cite{scissorbot2024,dexscrew2026,inhandassembly2026}.}

To tackle these challenges, recent studies have integrated tactile, force, and visuotactile feedback into dexterous manipulation policies~\cite{hato2025,touch2trace2026}. These approaches have unlocked complex manipulation capabilities, including apple peeling~\cite{tang2026humanlike}, nut--bolt fastening~\cite{dexscrew2026}, and handling fragile potato chips~\cite{fingereye2026}. Collectively, these advances underscore the critical role of contact feedback in coordinating finger motions and modulating interaction forces under dynamic contact states. However, contact-force perception is not universally the primary bottleneck in precise dexterous manipulation~\cite{gaze2021}. Consider cutting a pattern out of paper with scissors: the robot must continuously track the moving intersection of the blades relative to the target contour while adapting to real-time deformations and shifts in the paper's pose~\cite{scissorbot2024}. In such scenarios, alongside tactile feedback, task success also relies heavily on fine-grained visual perception of geometric and spatial relationships. Despite its importance, systematic discovery and exploitation of task-critical visual cues, such as tool--object alignment, contact geometry, and local deformation, remain  less explored.

To bridge this gap, we introduce \textbf{Focus4Dex}, a framework designed to identify which visual cues are task-critical and how to condition policy execution on these signals. As illustrated in~\cref{figs:teaser} and ~\cref{figs:pipeline}, our framework tackles these two questions through an active perception loop comprising three core components. First, Action-sensitive Visual Region (AVR) Discovery automatically identifies task-critical regions without manual annotations, providing scalable supervision for \emph{where to focus}. Second, an online AVR Localization Module is trained on these discovered regions to predict AVRs directly in real time during deployment. Third, turning to \emph{how to manipulate}, action-conditioned visual modulation encodes local visual features and region geometry from the predicted AVRs to modulate the action expert via gated adaptive RMS normalization (AdaRMS)~\cite{rmsnorm2019,dit2023}. Meanwhile, full-scene observations continue to follow the base policy's standard visual pathway without architectural disruption. By preserving the global visual representation alongside this localized conditioning, our policy effectively integrates scene-level context with fine-grained local focus for high-precision \mbox{dexterous manipulation. Our contributions are threefold:}
\begin{enumerate}
    \item \textbf{Focus4Dex.}  We introduce Focus4Dex, an action-guided framework that identifies and exploits task-critical visual regions for fine-grained dexterous tool use and deformable-object manipulation. Its three-part pipeline—AVR Discovery, the AVR Localization Module, and Action-conditioned Visual Modulation—connects offline action-sensitive region discovery with online localization and global-local visual conditioning \mbox{for precise action generation.}
    \item \textbf{Global-Local Visual Fusion.} We develop a visual conditioning mechanism that integrates global scene context, action-sensitive local visual features, and explicit region geometry into action generation \mbox{via gated suffix AdaRMS.}
    \item \textbf{Comprehensive Real-world Evaluation.} We evaluate our method on a bimanual 65-DoF dexterous robot across four challenging tasks spanning tool use and deformable object manipulation. Focus4Dex consistently outperforms all baselines on every task, achieving a 50.5\% mean success rate and surpassing the strongest baseline \mbox{by 12.5 percentage points.}
\end{enumerate}

\section{Related Work}
\label{sec:related_work}

\subsection{Precise Dexterous Manipulation}
Advances in dexterous hand hardware, scalable data collection interfaces~\cite{opentelevision2025,ace2025,bidex2025}, and policy learning~\cite{maniptrans2025} have unlocked diverse dexterous tool-use skills, such as bimanual paper cutting, apple peeling, and using tweezers, syringes, and spray bottles~\cite{dexcap2024,tang2026humanlike,dexumi2025,grasp2dexterity2026}. To coordinate contact-rich interactions, recent works, such as RotateIt, MoDE-VLA, HATO, DexScrew, and FingerEye, integrate force, torque, or visuotactile sensors to regulate contact forces during in-hand rotation, peeling, fastening, and delicate grasping~\cite{rotateit2023,tang2026humanlike,hato2025,dexscrew2026,fingereye2026}. For assembly and deformable objects, prior methods similarly rely on contact feedback or explicit state representations, including visuo-tactile refinement and neural fields~\cite{vtrefine2025,neuralfeels2024}, visually estimated object poses~\cite{inhandassembly2026}, tracked keypoints for bag knotting~\cite{dexknot2026}, and tactile sensing for cable tracing~\cite{touch2trace2026}. Despite these advances in tactile sensing and explicit state tracking, directly harnessing fine-grained visual perception for dexterous control has received far less attention. To explore its potential in precise manipulation, we introduce Focus4Dex to identify which observations are critical for task execution and how to integrate them into manipulation policies.

\subsection{Visual Region Selection}
{\looseness=1
External priors, such as human gaze and saliency~\cite{gaze2021,visarl2024,giava2026}, calibrated tool projections~\cite{palm2026}, and object proposals or segmentation~\cite{viola2023,groot2023,controlvla2025}, can localize task-relevant regions. However, they depend on extra hardware, fixed tool-frame assumptions, or auxiliary detectors. Action-driven methods instead leverage control objectives to guide visual focus. Active vision approaches learn viewpoint selection via demonstration or action-prediction objectives~\cite{avaloha2025,eyerobot2025,learningtolook2024}, yet require actuated cameras; coarse-to-fine policies progressively zoom into regions selected by Q-values, feature fields, or predicted action locations~\cite{c2fq2022,act3d2023,rvt22024}, but primarily target discrete end-effector pose estimation and struggle with high-DoF continuous finger articulation; and Seeker learns region selection via an action-supervised attention bottleneck~\cite{seeker2026}, which may constrain pre-trained VLM pathways. More directly, perturbation-based attribution methods such as the Interventional Significance Score (ISS) quantify regional importance by measuring action-prediction shifts under visual perturbations~\cite{iss2026}. Building on this attribution principle, Focus4Dex performs AVR Discovery by probing a reference policy with localized visual perturbations, then trains the AVR localization module on the teacher AVRs for real-time localization, avoiding human \mbox{annotations, external priors, and architectural intrusion.}}\par

\subsection{Local Feature Utilization}
{\looseness=1
Once task-relevant regions are identified, existing methods integrate local observations into control through distinct architectures. Dual-stage frameworks, such as gaze-based control and PALM, decouple coarse global reaching from local fine-grained manipulation~\cite{gaze2021,palm2026}, whereas multi-scale policies like EyeRobot and MResT fuse global and foveated features across spatial scales or update frequencies~\cite{eyerobot2025,mrest2023}. For direct spatial prediction, ARM, C2F-Q, and RVT-2 infer actions or poses from local crops, refined voxel grids, or re-rendered views~\cite{arm2022,c2fq2022,rvt22024}. Generative policies instead condition iterative action denoising on local representations: GIAVA and ControlVLA incorporate foveated or mask-derived tokens via cross-attention~\cite{giava2026,controlvla2025}, while Seeker applies FiLM modulation to combine crop features with box geometry for a diffusion U-Net~\cite{seeker2026}, but replacing the global view with the local crop may impair macro-level spatial reference and bimanual coordination. Rather than trading global context for local resolution, Focus4Dex retains both global and local representations and, guided by an empirical comparison of conditioning mechanisms, injects local visual features and region geometry into the action expert via \mbox{gated adaptive RMS normalization (AdaRMS)~\cite{rmsnorm2019,dit2023}.}}\par
\section{Method}
\label{sec:method}
\begin{figure*}
  \centering
  \includegraphics[width=\textwidth]{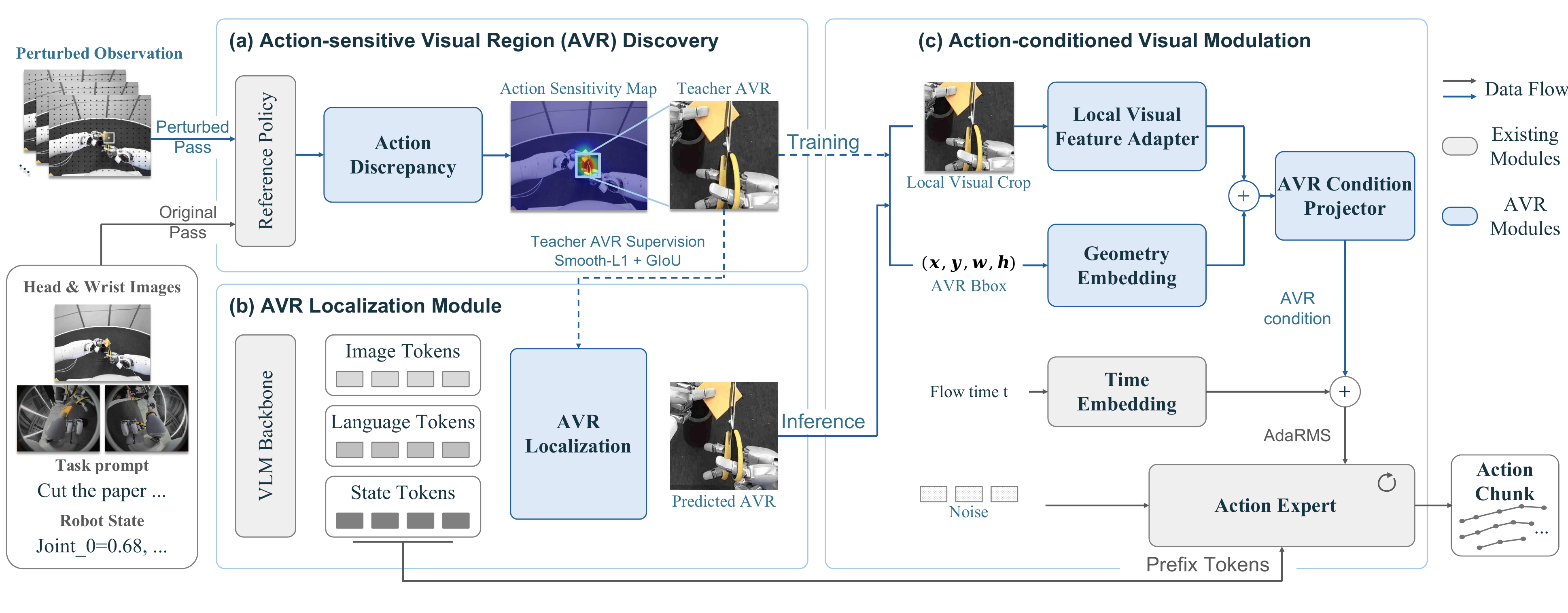}
  \caption{\textbf{Overview of the Focus4Dex Framework.} (a) \textbf{Action-sensitive Visual Region (AVR) Discovery} feeds clean and locally blurred images into a reference policy, converting action shifts into an Action Sensitivity Map to extract the teacher AVR. (b) \textbf{AVR Localization Module} processes the VLM global prefix (global views, language, and state) via an MLP head to regress online AVR predictions, supervised by the teacher AVRs from (a). (c) \textbf{Action-conditioned Visual Modulation} encodes the local visual crop via a local visual feature adapter and fuses the resulting features with a region geometry embedding through an AVR condition projector to modulate the action expert via gated suffix AdaRMS while reusing the cached global prefix. For training, we first train the reference policy in (a) without AVR inputs to mine offline teacher AVRs, which then condition the training of the action policy in (c). Next, with the policy frozen, teacher AVRs supervise AVR localization in (b), whose predictions condition the policy during deployment.}
  \label{figs:pipeline}
\end{figure*}

{
To address the visual challenges of \emph{where to focus} and \emph{how to manipulate} in fine-grained dexterous tool use and deformable-object manipulation, Focus4Dex introduces a three-part framework spanning policy training and online inference (\cref{figs:pipeline}): (1) \textbf{Action-sensitive Visual Region (AVR) Discovery} (\cref{sec:teacher}), which queries a reference policy offline under localized visual perturbations to automatically identify offline AVRs without manual spatial annotations; (2) an \textbf{AVR Localization Module} (\cref{sec:predictor}), supervised by the offline teacher AVRs to predict AVRs online from the global visual-language prefix without online perturbation overhead; and (3) \textbf{Action-conditioned Visual Modulation} (\cref{sec:adarms}), which fuses the local visual features and region geometry into a compact conditioning vector to modulate the action expert via gated adaptive RMS normalization (AdaRMS) while preserving global scene representations for precise dexterous manipulation.\par}

\subsection{Action-sensitive Visual Region Discovery}
\label{sec:teacher}

\begin{figure}
  \centering
  \subfloat[Paper Folding\label{fig:crop_fold}]{%
    \includegraphics[width=0.48\columnwidth]{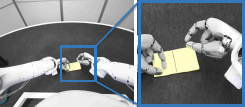}}%
  \hfill
  \subfloat[Paper Cutting\label{fig:crop_paper}]{%
    \includegraphics[width=0.48\columnwidth]{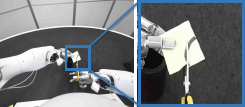}}\\[2pt]
  \subfloat[Paper Unfolding\label{fig:crop_unfold}]{%
    \includegraphics[width=0.48\columnwidth]{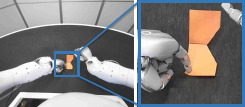}}%
  \hfill
  \subfloat[Thread Cutting\label{fig:crop_thread}]{%
    \includegraphics[width=0.48\columnwidth]{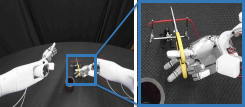}}
  \caption{\textbf{Examples of localized AVRs.} Each pair displays the full head-camera observation with its AVR bounding box (left) and the corresponding native-resolution local visual crop (right).}
  \label{fig:crop_examples}
\end{figure}

{
Full-scene images provide essential global context for bimanual coordination, while inevitably containing substantial task-irrelevant background. Capturing this global scene structure often comes at the expense of fine-grained visual details, which can shrink to only a few pixels after standard image downsampling and become further diluted during visual tokenization. However, these subtle cues—such as blade--contour intersections, paper edges, and contact states—are decisive for task success in precise dexterous manipulation. Therefore, to enhance policy performance on fine-grained manipulation, we seek to identify these task-critical local regions and supply them to the policy at high resolution for precise bimanual contact control. Rather than relying on external priors or altering the pretrained model pathway, we leverage an action-trained policy itself to automatically discover which visual \mbox{regions drive its control decisions.}\par}

Specifically, we first fine-tune a pretrained $\pi_{0.5}$ model on the task demonstrations and freeze its weights to serve as an offline reference policy. At demonstration step $t$, we quantify how localized visual perturbations shift its predicted actions (\cref{figs:pipeline}(a)). Given the clean head-camera image $I_t^{\mathrm{head}}$, we evaluate the action discrepancy of the reference policy on the clean observation and on a grid of overlapping $64\times64$ Gaussian-blurred windows (a $16\times16$ grid yielding 256 perturbed views $I_t^{(j)}$), while keeping the wrist images, proprioceptive state, and language instruction intact. To eliminate stochasticity from flow-matching sampling, both clean and perturbed passes share identical initial noise over 10 denoising steps. Let $A^{(0)}, A^{(j)} \in \mathbb{R}^{50\times 65}$ denote the normalized action chunks predicted from the clean observation and the $j$-th perturbed observation, respectively. We measure the local action sensitivity $d_j$ over the first $K=20$ steps and the $|\mathcal{I}_{\mathrm{act}}|=65$ active DoFs as
\begin{equation}
 d_j = \sqrt{\frac{1}{K\,|\mathcal{I}_{\mathrm{act}}|}\sum_{k=1}^{K}\sum_{a\in\mathcal{I}_{\mathrm{act}}}
 \left(A^{(j)}_{ka}-A^{(0)}_{ka}\right)^2}.
 \label{eq:sensitivity}
\end{equation}
{
    A larger discrepancy $d_j$ indicates stronger reliance of the reference policy on the visual evidence occluded at location $j$. To convert these window-level scores into a continuous spatial field, we aggregate the discrepancy scores $d_j$ weighted by each window's soft mask, followed by Gaussian smoothing and min--max normalization to produce an Action Sensitivity Map $\mathcal{S}_t$. From each smoothed map $\mathcal{S}_t$, we extract the connected high-response component containing the global peak to obtain its center and spatial extent, which are then squared, expanded, and clamped to the image boundaries to yield the teacher AVR bounding box $b_t$. Mapping $b_t$ back onto the native-resolution head image $I_t^{\mathrm{head}}$ to extract the corresponding local visual crop $C_t$ preserves fine-grained geometric details that would otherwise be lost during global resizing. Representative discovered AVRs across all four tasks are shown in \cref{fig:crop_examples}, and both $b_t$ and $C_t$ serve as offline signals for region localization and action policy conditioning.\par}

\subsection{AVR Localization Module}
\label{sec:predictor}
{\looseness=1
Although perturbation-based discovery yields reliable AVR supervision offline, executing dense perturbation sweeps online requires hundreds of forward passes per control step, incurring prohibitive latency. To determine \emph{where to focus} in real time without online perturbation queries, we introduce a lightweight AVR Localization Module (\cref{figs:pipeline}(b)). It operates directly on the VLM global prefix tokens, applying LayerNorm, learned attention pooling, and a compact MLP head to regress a normalized in-frame bounding box $\hat{b}_t$. Because the global VLM representations continue to adapt during action policy training (\cref{sec:adarms}), we train this module in a subsequent stage once the AVR-conditioned action policy has converged. Specifically, with all other policy components kept frozen, we optimize only the AVR Localization Module using a combination of Smooth-L1 and generalized IoU ($\operatorname{GIoU}$) losses supervised by the teacher AVR bounding boxes $b_t$. At inference time, the predicted box $\hat{b}_t$ is mapped back onto the native-resolution head-camera image $I_t^{\mathrm{head}}$ to extract the online local visual \mbox{crop $\hat{C}_t$ in a single forward pass.}\par}

\subsection{Action-conditioned Visual Modulation}
\label{sec:adarms}
Given a localized AVR, a key design question is how the policy can effectively integrate its fine-grained visual cues to execute precise manipulation. As shown in \cref{figs:pipeline}(c), both the global camera views and the local visual crop ($C_t$ during training, $\hat{C}_t$ at deployment) are encoded by the VLM's shared, trainable SigLIP vision encoder, aligning global and local features within a unified representation space under the action-learning objective. To form a compact local descriptor, a local visual feature adapter spatially downsamples the crop tokens, projects them to the hidden dimension of the action expert, and mean-pools them into pooled local visual features $\bar{z}_t$. Because cropping and resizing discard the region's absolute location and scale within the workspace, the geometry embedding module explicitly encodes the normalized center and spatial extent of the AVR box ($b_t$ during training, $\hat{b}_t$ at deployment) alongside a validity indicator. An AVR condition projector $f_{\mathrm{AVR}}$ then concatenates and maps $\bar{z}_t$ and this geometric embedding into a unified vector, whose RMS is capped at one. For a valid local visual crop at flow-matching step $\tau \in [0, 1]$, we construct the composite conditioning vector $c_{t,\tau}$ by augmenting the flow-time \mbox{embedding $e(\tau)$ through a learnable scalar gate $\gamma$:}
\begin{equation}
 c_{t,\tau}=e(\tau)+\tanh(\gamma)\,
 f_{\mathrm{AVR}}(\bar{z}_t,b_t),
 \label{eq:condition}
\end{equation}
{\looseness=1
where $b_t$ is replaced by $\hat{b}_t$ during online deployment. Rather than appending local tokens to the VLM sequence, $c_{t,\tau}$ modulates the 18-block Transformer action expert via adaptive RMS normalization (AdaRMS)~\cite{rmsnorm2019,dit2023}. Within each Transformer block, dedicated linear projections map $c_{t,\tau}$ to adaptive scale and shift parameters for pre-layer RMS normalization~\cite{rmsnorm2019}, as well as element-wise gates for the attention and feed-forward residual branches~\cite{dit2023}; an additional projection modulates the final normalization layer prior to velocity prediction. This design injects fine-grained contact and alignment cues throughout the entire action-denoising suffix while \mbox{leaving the global VLM prefix intact.}\par}

To train the conditioned policy, we jointly optimize the VLM backbone 
(including the shared SigLIP encoder), the local conditioning modules, and the 
action expert under the conditional flow-matching objective using offline local visual crops. We initialize $\gamma = 0$ so that conditioning starts smoothly from the base policy's behavior, and apply crop dropout which substitutes a learned null condition for randomly dropped or invalid crops to promote robustness.

At deployment, online localization (\cref{sec:predictor}) and action 
modulation operate together in a shared-prefix loop: (1) the VLM encodes the 
global views, language prompt, and robot state once, caching its prefix keys and values; (2) the AVR localization module predicts $\hat{b}_t$ from the VLM prefix tokens to extract the native-resolution local visual crop $\hat{C}_t$; (3) the local branch computes $c_{t,\tau}$ and the action expert generates an action chunk while reusing the cached global prefix. Compared with training, this adds only a single lightweight bounding-box regression pass. Further method details are provided in \cref{sec:method_details} of the supplementary material.
\section{Experiments}
\label{sec:experiments}
We evaluate Focus4Dex on a real-world 65-DoF bimanual dexterous humanoid platform across four dexterous tool-use and deformable-object manipulation tasks. Specifically, our empirical study investigates three core questions:

(1) In \cref{sec:main_exp}, does incorporating fine-grained local visual details benefit dexterous manipulation?

(2) In \cref{sec:ablation_exp}, how effective are our \emph{where to} \emph{focus} and \emph{how to manipulate} strategies?

(3) In \cref{sec:localization_analysis}, what is the computational cost of enabling fine-grained visual focus during deployment?

\begin{figure*}
\centering
\begingroup
\edef\savedtaskcounter{\arabic{enumiv}}
\setcounter{enumiv}{0}
\newcommand{\tasksequencerow}[1]{%
  \stepcounter{enumiv}%
  \noindent\raisebox{-0.5\height}{\makebox[0.03\textwidth][c]{\small(\alph{enumiv})}}%
  \hspace{0.01\textwidth}%
  \raisebox{-0.5\height}{\includegraphics[width=0.94\textwidth]{#1}}%
  \par\vspace{2.5pt}}
\tasksequencerow{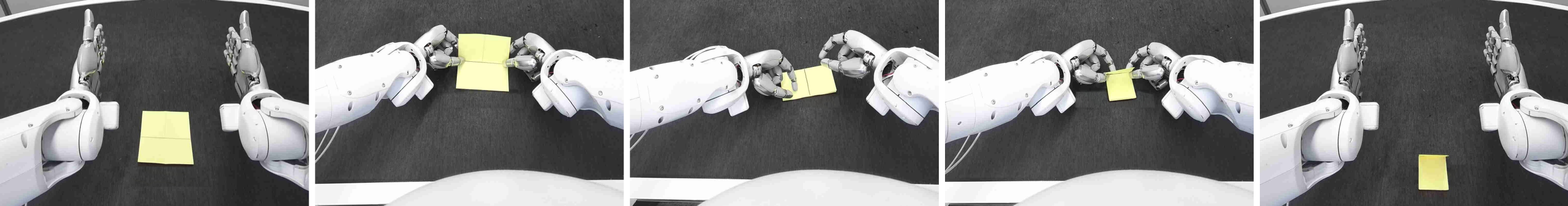}
\tasksequencerow{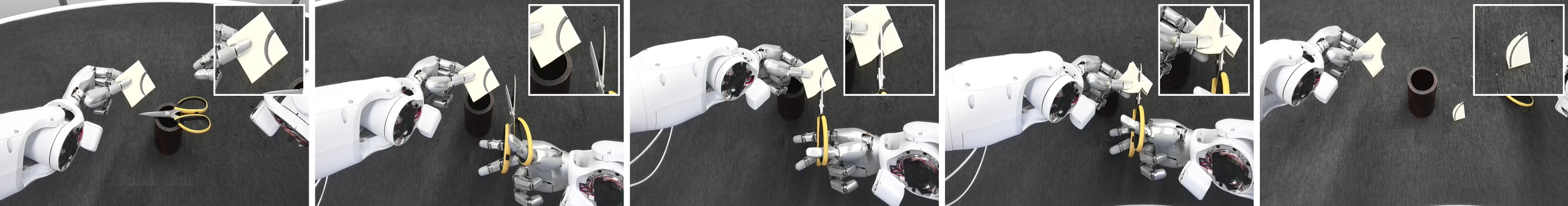}
\tasksequencerow{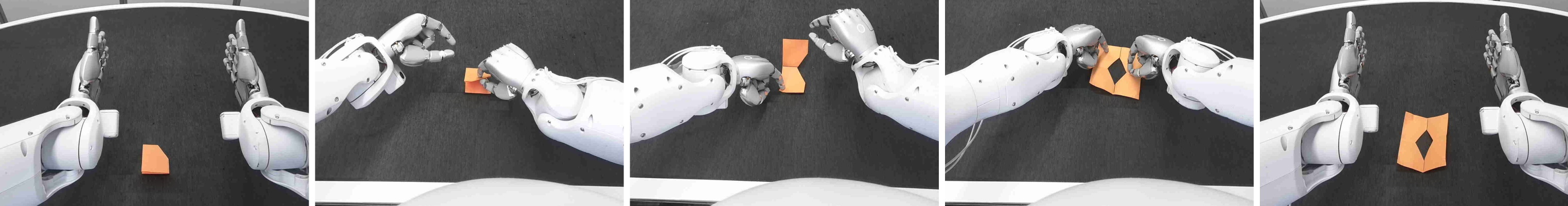}
\tasksequencerow{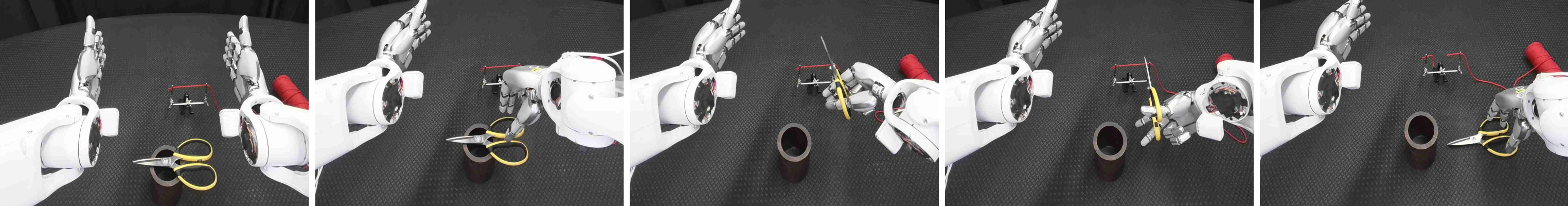}
\setcounter{enumiv}{\savedtaskcounter}
\endgroup
\caption{\textbf{Demonstrations of the Four Evaluated Tasks.} Each row presents five chronological keyframes captured by a head-mounted camera during a single demonstration of (a) paper folding, (b) paper cutting, (c) paper unfolding, and (d) thread cutting.}
\label{fig:task_sequences}
\end{figure*}

\subsection{Experimental Setup}
\label{sec:exp_setup}

\textbf{Hardware Platform.}
All experiments are conducted on the Sharpa North humanoid platform~\cite{tang2026humanlike}, a 65-DoF bimanual system comprising two 7-DoF arms, two 22-DoF Sharpa W01 anthropomorphic dexterous hands, a 2-DoF neck, a 3-DoF upper body, and a 2-DoF waist. Visual observations are provided by two head-mounted RGB cameras and two wrist-mounted fisheye cameras.

\textbf{Benchmark Tasks.}
We evaluate performance across four contact-rich real-world tasks (\cref{fig:task_sequences}) spanning articulated tool manipulation and deformable sheet handling. Paper folding, cutting, and unfolding demand continuous bimanual coordination to stabilize and reshape compliant sheets, whereas paper cutting and thread cutting require precise in-hand actuation of scissors relative to thin targets:
\begin{itemize}[leftmargin=*,nosep,topsep=1.5pt]
    \item \textbf{Paper Folding:} The robot coordinates both dexterous hands to execute two successive folds on a flat sheet of paper.
    \item \textbf{Paper Cutting:} The robot stabilizes a sheet of paper with its left hand while actuating scissors with its right hand to cut along a marked straight or curved black contour and separate the designated portion.
    \item \textbf{Paper Unfolding:} Starting from a sheet folded in half twice, the robot coordinates both hands to separate and flatten both folds in reverse order until the sheet is restored to its unfolded state.
    \item \textbf{Thread Cutting:} The robot grasps scissors with its right hand, approaches a thin thread secured to a stand, aligns the blades around the thread, and closes them to sever it.
\end{itemize}

\textbf{Implementation Details and Baselines.}
{\looseness=1
We initialize Focus4Dex from the pretrained $\pi_{0.5}$ model~\cite{pi05} and fine-tune it on task demonstrations for two epochs with a batch size of 16. At each replanning step, the VLM prefix encodes the right head-camera view and both wrist-camera views alongside the language prompt and 65-DoF proprioceptive state, while the online AVR localization module (\cref{sec:predictor}) predicts $\hat{b}_t$ from the VLM prefix tokens to extract a native-resolution local visual crop $\hat{C}_t$. Both global images and local visual crops are resized to $224 \times 224$ and encoded via the shared SigLIP backbone. The 18-block Transformer action expert predicts 50-step, 65-DoF action chunks using 10 flow-matching denoising steps and executes the first 20 steps before replanning. We compare Focus4Dex against five baselines: \textbf{LingBot-VA}~\cite{lingbot}, a causal video-action world model unifying visual prediction and control; \textbf{OpenWAM}~\cite{wang2026openwam}, a world-action model framework that couples video prediction with action generation; \textbf{DM0.5}~\cite{dm05}, a multimodal manipulation model with a memory design; \textbf{$\pi_0$}~\cite{pi0}, a flow-matching VLA model pairing a pretrained VLM with an action expert; and \textbf{$\pi_{0.5}$}~\cite{pi05}, our direct base VLA policy \mbox{pretrained for open-world generalization.}\par}

\textbf{Evaluation Protocol.}
{\looseness=1
We report task success rate (SR), defined as the percentage of rollouts satisfying precise, task-specific criteria: executing two successive half-folds (\emph{Paper Folding}); achieving at least 70\% trajectory overlap along the marked black line and completely separating the designated portion (\emph{Paper Cutting}); fully reversing both folds to restore the sheet to its initial configuration (\emph{Paper Unfolding}); and completely severing the thread secured to the stand (\emph{Thread Cutting}). All methods in the main comparison are evaluated under identical demonstration datasets and initial-scene distributions. Specifically, each method in \cref{tab:baseline} is assessed over 50 trials per task, whereas each ablation variant in \cref{tab:ablations} is evaluated over 20 trials per task. Tables maintain a uniform task order and report the \mbox{unweighted mean across the four tasks.}\par}

\begin{table*}[!t]
\centering\small
\caption{\textbf{Main Real-World Comparison.} Task success rates (\%) over 50 trials per task and method on the 65-DoF bimanual platform. Mean denotes the unweighted average across tasks; bold marks the highest observed success rate in each column.}
\label{tab:baseline}
\begin{tabular}{lccccc}
\toprule
Policy & Paper Folding & Paper Cutting & Paper Unfolding & Thread Cutting & Mean\\
\midrule
LingBot-VA~\cite{lingbot} & 16\% & 0\% & 10\% & 42\% & 17.0\%\\
OpenWAM~\cite{wang2026openwam} & 18\% & 0\% & 12\% & 48\% & 19.5\%\\
DM0.5~\cite{dm05} & 42\% & 0\% & 26\% & 12\% & 20.0\%\\
$\pi_0$~\cite{pi0} & 60\% & 22\% & 12\% & 34\% & 32.0\%\\
$\pi_{0.5}$~\cite{pi05} & 62\% & 36\% & 18\% & 36\% & 38.0\%\\
\midrule
\textbf{Focus4Dex (Ours)} & \textbf{76\%} & \textbf{42\%} & \textbf{30\%} & \textbf{54\%} & \textbf{50.5\%}\\
\bottomrule
\end{tabular}
\end{table*}

\begin{figure}
\centering
\begingroup
\setlength{\unitlength}{\linewidth}
\begin{picture}(1,0.48764)
  \put(0.12,0){\includegraphics[width=0.88\linewidth]{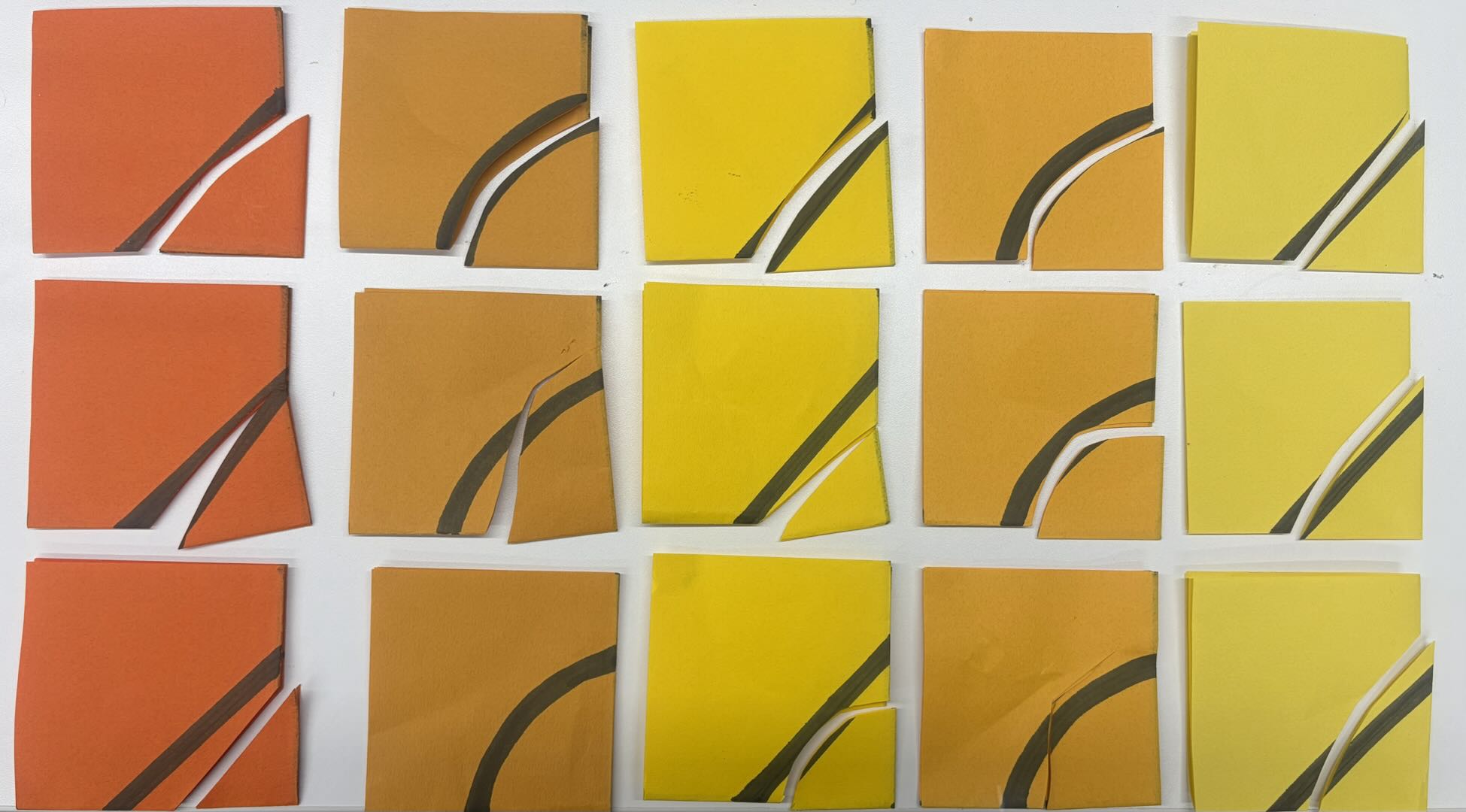}}
  \put(0.10,0.40410){\makebox(0,0)[r]{\scriptsize\textbf{Ours}}}
  \put(0.10,0.23930){\makebox(0,0)[r]{\scriptsize $\pi_{0.5}$}}
  \put(0.10,0.07676){\makebox(0,0)[r]{\scriptsize $\pi_0$}}
\end{picture}
\endgroup
\caption{\textbf{Qualitative Comparison of Paper-Cutting Outcomes.} Rows show representative cut specimens from Focus4Dex (Ours), $\pi_{0.5}$, and $\pi_0$ across straight and curved target paths (black lines).}
\label{fig:cutting_comparison}
\end{figure}

\subsection{Main Results (Q1)}
\label{sec:main_exp}

\textbf{Fine-grained local visual details boost dexterous performance.}
{
As reported in \cref{tab:baseline}, Focus4Dex achieves the highest success rate across all four evaluated tasks, attaining an overall mean success rate of \textbf{50.5\%} across 200 real-world trials—an improvement of \textbf{+12.5 percentage points} over the strongest baseline, $\pi_{0.5}$ (38.0\%). Under standard full-scene observations downsampled to $224 \times 224$, fine task-critical structures—such as thin threads, printed ink contours, and subtle paper creases—are compressed to only a few pixels, severely degrading spatial perception. Because Focus4Dex shares the exact same pretrained backbone, demonstration datasets, global workspace views, 65-DoF action expert capacity, and training procedure as $\pi_{0.5}$, this controlled comparison directly isolates the efficacy of our online action-guided local focus in overcoming perceptual limitations in fine-grained \mbox{bimanual manipulation tasks.}\par}

Notably, while Focus4Dex outperforms $\pi_{0.5}$ by at least 12 percentage points on thread cutting (+18 points, 54\% vs.\ 36\%), paper folding (+14 points, 76\% vs.\ 62\%), and paper unfolding (+12 points, 30\% vs.\ 18\%), its binary success-rate gain on paper cutting is a more modest +6 percentage points (42\% vs.\ 36\%). Nevertheless, binary completion metrics alone do not fully reflect continuous execution fidelity. As evidenced by the straight and curved specimens in \cref{fig:cutting_comparison}, Focus4Dex tracks the printed target contours with higher spatial precision throughout execution, yielding clean cut edges that align noticeably more closely with the reference lines than those produced by $\pi_{0.5}$ and $\pi_0$.

On paper unfolding, by contrast, DM0.5 achieves a 26\% success rate, outperforming $\pi_{0.5}$ (18\%), OpenWAM (12\%), and $\pi_0$ (12\%) among the baselines. Because separating stacked paper layers with dexterous fingertips rarely succeeds on a single attempt, human demonstrations naturally contain frequent recovery and retry maneuvers after initial slips. DM0.5's memory design may help it learn these temporal retry patterns from the demonstrations. Nevertheless, Focus4Dex still secures the highest unfolding success rate (\textbf{30\%}) by jointly integrating global context with high-resolution local details, demonstrating that finer visual perception may enable more accurate physical execution.
\begin{table*}
\centering
\small
\setlength{\tabcolsep}{6pt}
\renewcommand{\arraystretch}{1.08}
\caption{\textbf{Ablation Study on Visual Region Supervision and Conditioning Architectures.} Task success rates (\%) over 20 trials per task. All six configurations predict local regions online at deployment and supply identical explicit region geometry encodings; the five AVR-based routes share matched training data, encoders, and optimization steps. Bold marks each column's maximum.}
\label{tab:ablations}
\begin{tabular*}{\textwidth}{@{\extracolsep{\fill}}lccccc@{}}
\toprule
Configuration & \shortstack{Paper\\Folding} & \shortstack{Paper\\Cutting} & \shortstack{Paper\\Unfolding} & \shortstack{Thread\\Cutting} & Average\\
\midrule
Prefix Concatenation (predicted AVR) & 65\% & 10\% & 10\% & 15\% & 25.00\%\\
Prefix Resampler (predicted AVR) & 70\% & 20\% & 5\% & 40\% & 33.75\%\\
Suffix Tokens (predicted AVR) & 85\% & 20\% & 5\% & 45\% & 38.75\%\\
Suffix Cross-Attention (predicted AVR) & 80\% & 0\% & 30\% & 70\% & 45.00\%\\
\midrule
Suffix AdaRMS (object-centric region) & \textbf{90\%} & 25\% & 30\% & 65\% & 52.50\%\\
\textbf{Suffix AdaRMS (predicted AVR; Ours)} & 85\% & \textbf{35\%} & \textbf{45\%} & \textbf{75\%} & \textbf{60.00\%}\\
\bottomrule
\end{tabular*}
\end{table*}

\begin{figure}
\centering
\begingroup
\newcommand{\localcroprow}[3]{%
  \noindent\raisebox{-0.5\height}{\makebox[0.055\linewidth][c]{%
    \rotatebox{90}{\scriptsize\textcolor[gray]{#1}{#2}}}}%
  \hfill\raisebox{-0.5\height}{\includegraphics[width=0.93\linewidth]{#3}}%
  \par\vspace{2pt}}
\subfloat[Paper Folding]{\begin{minipage}{0.49\linewidth}
\localcroprow{0.5}{Object-Centric}{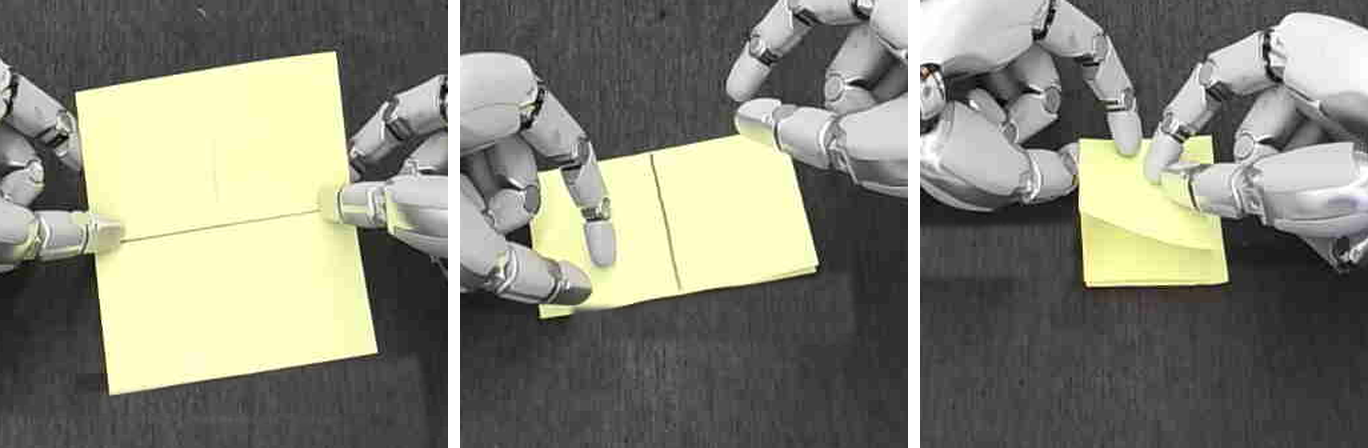}
\localcroprow{0}{AVR}{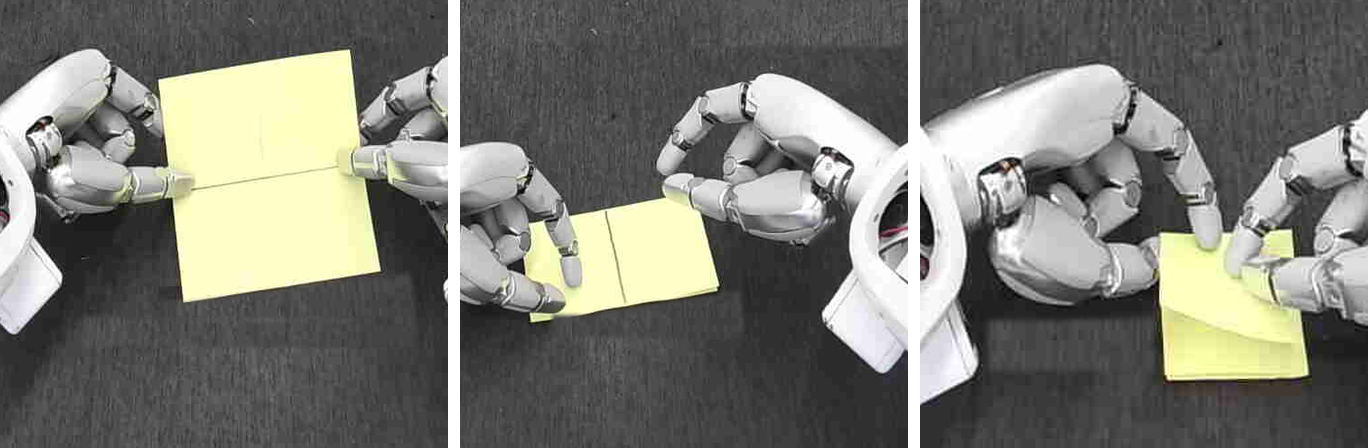}
\end{minipage}}\hfill
\subfloat[Paper Cutting]{\begin{minipage}{0.49\linewidth}
\localcroprow{0.5}{Object-Centric}{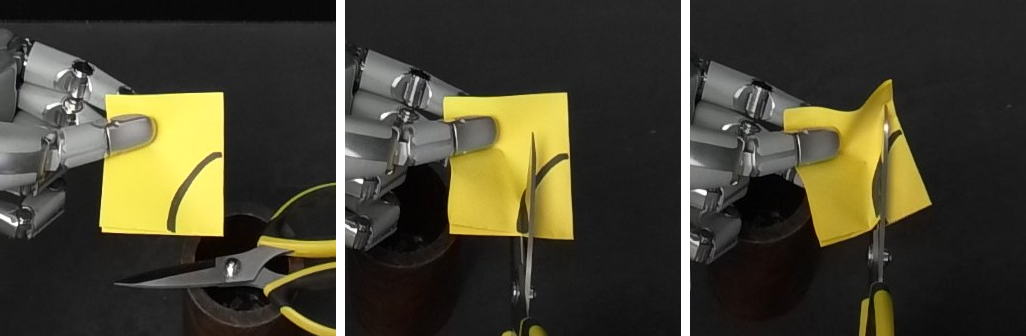}
\localcroprow{0}{AVR}{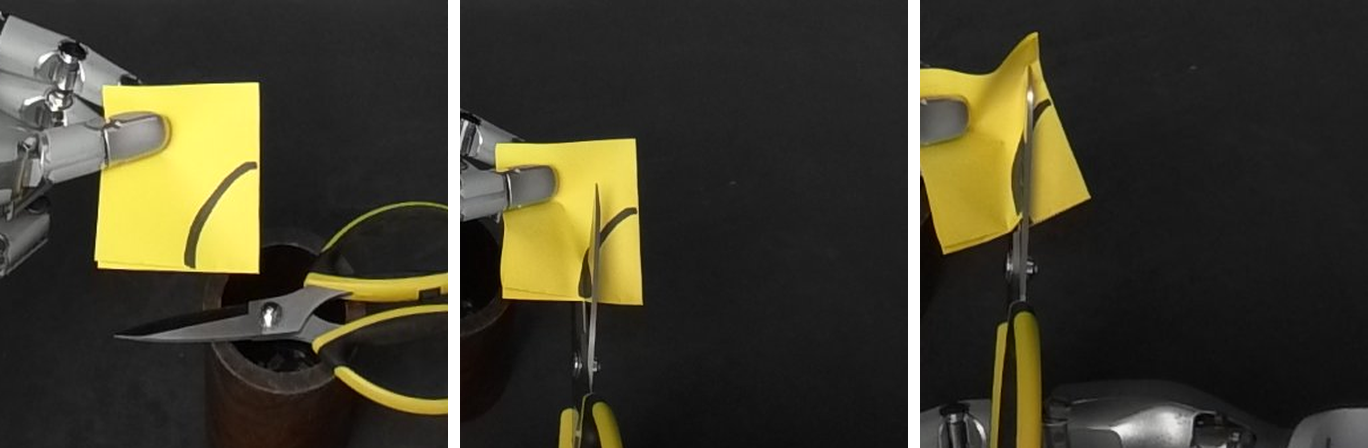}
\end{minipage}}\par\vspace{2.5pt}
\subfloat[Paper Unfolding]{\begin{minipage}{0.49\linewidth}
\localcroprow{0.5}{Object-Centric}{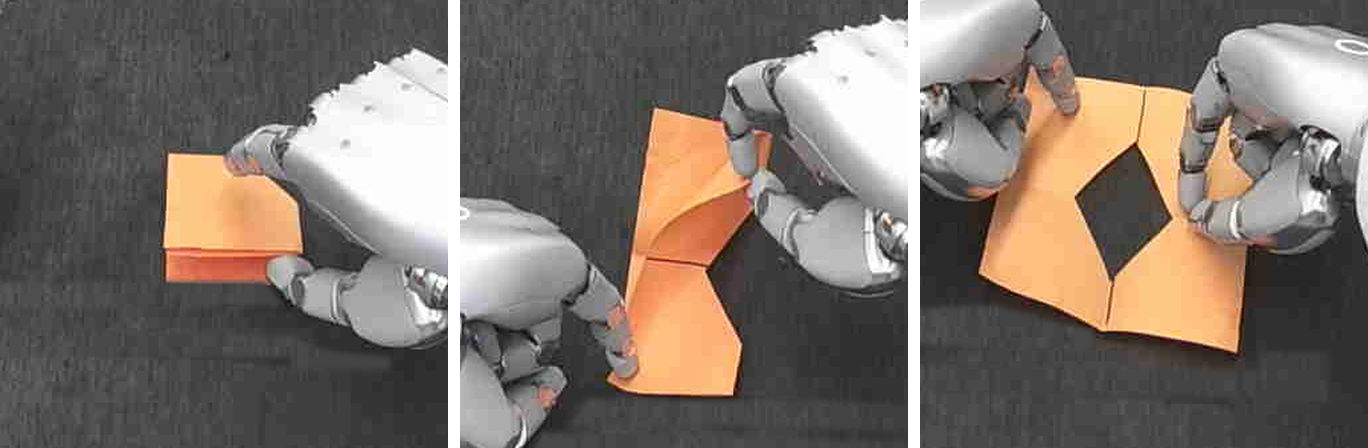}
\localcroprow{0}{AVR}{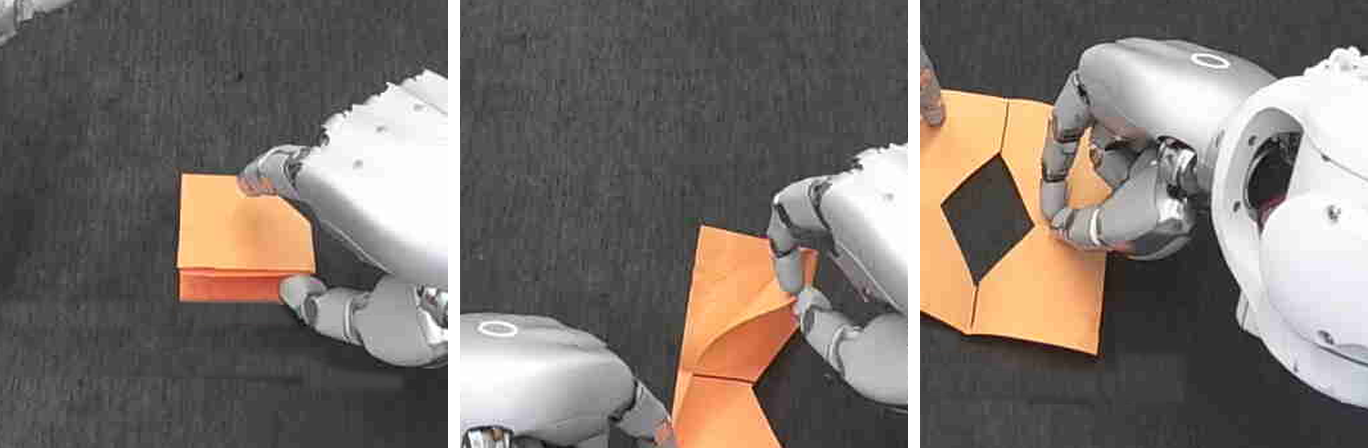}
\end{minipage}}\hfill
\subfloat[Thread Cutting]{\begin{minipage}{0.49\linewidth}
\localcroprow{0.5}{Object-Centric}{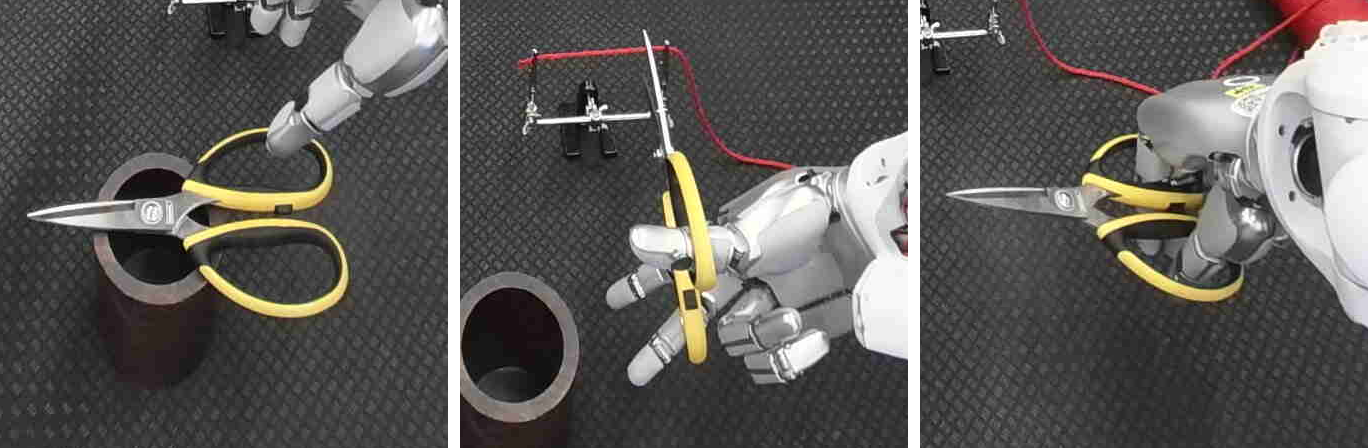}
\localcroprow{0}{AVR}{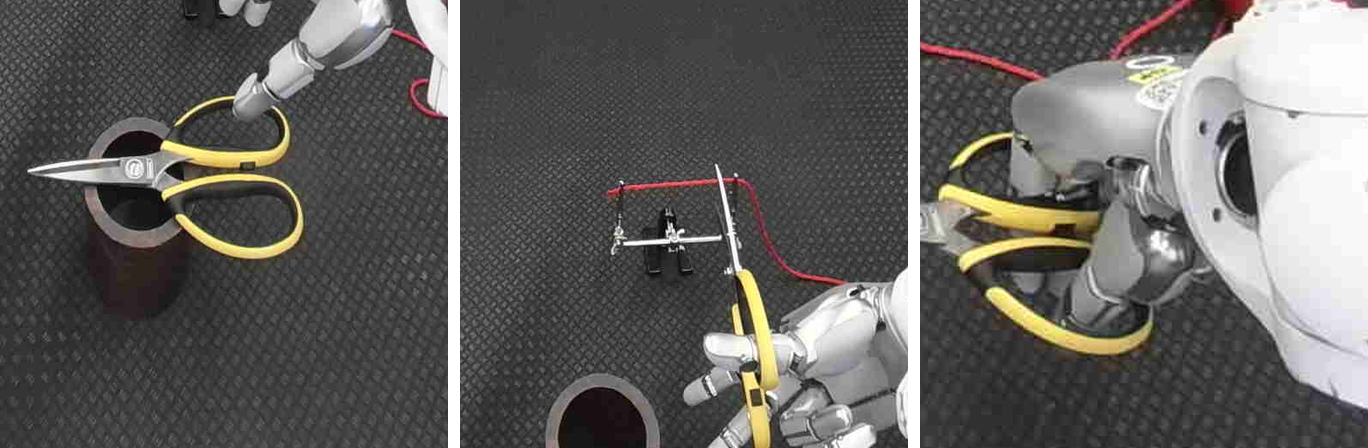}
\end{minipage}}
\endgroup
\caption{\textbf{Comparison of Local Visual Crops across Four Tasks.} Panels show (a) paper folding, (b) paper cutting, (c) paper unfolding, and (d) thread cutting across three chronological frames from one demonstration. Top rows show object-centric crops derived from SAM3.1 segmentation priors, while bottom rows show our discovered AVRs concentrating on task-critical interaction zones.}
\label{fig:crop_annotation_comparison}
\end{figure}

\subsection{Ablation Studies (Q2)}
\label{sec:ablation_exp}
In this section, we conduct ablation studies across all four manipulation tasks in~\cref{tab:ablations} to investigate the contributions of Focus4Dex's two core designs: \emph{where to focus} via AVR Discovery and \emph{how to manipulate} through Action-conditioned Visual Modulation with gated suffix AdaRMS.

\textbf{Where to Focus: AVRs vs.\ Object-Centric Priors.}
We first compare teacher AVRs obtained by offline perturbation (\cref{sec:teacher}) against object-centric region priors~\cite{viola2023,groot2023,controlvla2025} under identical gated suffix AdaRMS modulation (\cref{tab:ablations}, bottom 
two rows). In the object-centric variant, boxes curated from SAM3.1 segmentation masks serve as both conditioning input and localization module supervision; at deployment, both policies predict target crops online via their respective heads. Replacing object-centric regions with discovered AVRs raises mean success from 52.50\% to \textbf{60.00\%} (+7.50 percentage points), improving paper cutting (+10 points, 35\% vs.\ 25\%), unfolding (+15 points, 45\% vs.\ 30\%), and thread cutting (+10 points, 75\% vs.\ 65\%), while object-centric priors retain a one-trial lead on folding (90\% vs.\ 85\%). \Cref{fig:crop_annotation_comparison} shows representative examples comparing the two types of crops across all four tasks: whereas object-centric crops tend to cover the broader target object, our discovered AVRs attend more closely to the local contact zones where manipulation actively occurs.

{\looseness=1
\textbf{How to Manipulate: Conditioning Architecture.} Using the same teacher AVRs for supervision, we benchmark five routes for injecting local visual features and region geometry into action generation (\cref{tab:ablations}):
(1) \textbf{Prefix Concatenation} appends crop tokens directly to the VLM sequence;
(2) \textbf{Prefix Resampler} compresses crop tokens via a learned resampler before VLM insertion;
(3) \textbf{Suffix Tokens} appends projected local tokens to the action sequence inside the action expert;
(4) \textbf{Suffix Cross-Attention} adds cross-attention layers in the action expert over local visual feature tokens; and
(5) \textbf{Suffix AdaRMS (Ours)} incorporates projected local features into the flow-time conditioning to modulate the action expert through gated adaptive RMS normalization (\cref{sec:adarms}). All five routes share the same training data, trainable SigLIP encoders, and optimization steps, with each localization head trained on its route's converged VLM prefix. Mean success spans 35 percentage points across architectures, rising from prefix concatenation (25.00\%) to gated suffix AdaRMS (\textbf{60.00\%}). Conditioning the action suffix outperforms modifying the VLM prefix (25.00\% and 33.75\%), as inserting crop tokens into the VLM sequence may disrupt pretrained multimodal pathways. Within the action expert, gated AdaRMS surpasses both suffix tokens (38.75\%) and suffix cross-attention (45.00\%): suffix tokens can similarly perturb the attention sequence, while additional cross-attention layers might introduce instability related \mbox{to global-context loss or representation mismatch.}\par}

\subsection{Inference Efficiency and Latency Analysis (Q3)}
\label{sec:localization_analysis}
While incorporating local visual features introduces additional visual inputs, Focus4Dex is designed to keep the impact on inference speed minimal. During inference, each observation requires only a single global VLM forward pass, which simultaneously supplies localization features and populates a shared KV cache reused across all denoising steps; because local visual features and region geometry condition the action expert through gated AdaRMS without adding attention tokens, the extra computation over $\pi_{0.5}$ is strictly confined to the AVR Localization Module, local visual crop encoding, and lightweight projections. Empirically, we benchmark our optimized inference path on a single NVIDIA GeForce RTX 4090 (batch size 1, 10 denoising steps; 100 inference calls), achieving a mean latency of \textbf{152.7\,ms} versus \textbf{131.2\,ms} for matched $\pi_{0.5}$, confirming a modest overhead for real-time execution. Finally, while both models currently operate synchronously, our architecture can also integrate with asynchronous execution frameworks such as Real-Time Chunking (RTC)~\cite{black2025realtime} without architectural modifications. See \cref{sec:experimental_details,sec:additional_results} in the supplementary material for further experiment details and results.
\section{Conclusion}
\label{sec:conclusion}
In this paper, we presented Focus4Dex, an effective framework for
fine-grained visual perception in precise dexterous manipulation.
Unifying task-critical visual discovery with local action conditioning,
our method addresses both \emph{where to focus} and \emph{how to
manipulate} in these tasks. By using AVR Discovery to supervise the AVR
Localization Module and applying action-conditioned visual modulation
through gated AdaRMS, Focus4Dex integrates fine-grained spatial evidence
with global scene context. Evaluations across four challenging real-world
tasks demonstrate that Focus4Dex outperforms existing baselines with a
50.5\% mean success rate, validating the effectiveness of AVR
localization, action-conditioned visual modulation, and real-time inference with low latency overhead.

\clearpage
\section{LLM Usage Statement}
OpenAI Codex assisted with restructuring and editing manuscript text, checking literature and LaTeX sources, generating internal review feedback, and revising labels in an existing method diagram. Google Gemini was also used to polish the language of the manuscript. This assistance did not generate experimental measurements. The authors take full responsibility for all scientific claims, references, and content presented in the final manuscript.
{\small
\bibliographystyle{focus4dex_refs}
\bibliography{references}

\begin{thebibliography}{44}
\providecommand{\natexlab}[1]{#1}
\providecommand{\url}[1]{\texttt{#1}}
\expandafter\ifx\csname urlstyle\endcsname\relax
  \providecommand{\doi}[1]{doi: #1}\else
  \providecommand{\doi}{doi: \begingroup \urlstyle{rm}\Url}\fi

\bibitem[Black et~al.(2025{\natexlab{a}})Black, Brown, Darpinian, Dhabalia,
  Driess, Esmail, Equi, Finn, Fusai, Galliker, Ghosh, Groom, Hausman, Ichter,
  Jakubczak, Jones, Ke, LeBlanc, Levine, Li-Bell, Mothukuri, Nair, Pertsch,
  Ren, Shi, Smith, Springenberg, Stachowicz, Tanner, Vuong, Walke, Walling,
  Wang, Yu, and Zhilinsky]{pi05}
K.~Black, N.~Brown, J.~Darpinian, K.~Dhabalia, D.~Driess, A.~Esmail, M.~R.
  Equi, C.~Finn, N.~Fusai, M.~Y. Galliker, D.~Ghosh, L.~Groom, K.~Hausman,
  B.~Ichter, S.~Jakubczak, T.~Jones, L.~Ke, D.~LeBlanc, S.~Levine, A.~Li-Bell,
  M.~Mothukuri, S.~Nair, K.~Pertsch, A.~Z. Ren, L.~X. Shi, L.~Smith, J.~T.
  Springenberg, K.~Stachowicz, J.~Tanner, Q.~Vuong, H.~Walke, A.~Walling,
  H.~Wang, L.~Yu, and U.~Zhilinsky.
\newblock $\pi_{0.5}$: a vision-language-action model with open-world
  generalization.
\newblock In \emph{Conference on Robot Learning (CoRL)}, 2025.

\bibitem[Black et~al.(2025{\natexlab{b}})Black, Brown, Driess, Esmail, Equi,
  Finn, Fusai, Groom, Hausman, Ichter, Jakubczak, Jones, Ke, Levine, Li-Bell,
  Mothukuri, Nair, Pertsch, Shi, Smith, Tanner, Vuong, Walling, Wang, and
  Zhilinsky]{pi0}
K.~Black, N.~Brown, D.~Driess, A.~Esmail, M.~R. Equi, C.~Finn, N.~Fusai,
  L.~Groom, K.~Hausman, B.~Ichter, S.~Jakubczak, T.~Jones, L.~Ke, S.~Levine,
  A.~Li-Bell, M.~Mothukuri, S.~Nair, K.~Pertsch, L.~X. Shi, L.~Smith,
  J.~Tanner, Q.~Vuong, A.~Walling, H.~Wang, and U.~Zhilinsky.
\newblock $\pi_0$: A vision-language-action flow model for general robot
  control.
\newblock In \emph{Robotics: Science and Systems (RSS)}, 2025.

\bibitem[Black et~al.(2025{\natexlab{c}})Black, Galliker, and
  Levine]{black2025realtime}
K.~Black, M.~Y. Galliker, and S.~Levine.
\newblock Real-time execution of action chunking flow policies.
\newblock In \emph{Advances in Neural Information Processing Systems
  (NeurIPS)}, 2025.

\bibitem[Cheng et~al.(2025)Cheng, Li, Yang, Yang, and Wang]{opentelevision2025}
X.~Cheng, J.~Li, S.~Yang, G.~Yang, and X.~Wang.
\newblock {Open-TeleVision}: Teleoperation with immersive active visual
  feedback.
\newblock In \emph{Conference on Robot Learning (CoRL)}, 2024.

\bibitem[Chuang et~al.(2025)Chuang, Lee, Gao, Naddaf-Sh, and
  Soltani]{avaloha2025}
I.~Chuang, A.~Lee, D.~Gao, M.-M. Naddaf-Sh, and I.~Soltani.
\newblock Active vision might be all you need: Exploring active vision in
  bimanual robotic manipulation.
\newblock In \emph{IEEE International Conference on Robotics and Automation
  (ICRA)}, 2025.

\bibitem[Chuang et~al.(2026)Chuang, Zou, Lee, Gao, and Soltani]{giava2026}
I.~Chuang, J.~Zou, A.~Lee, D.~Gao, and I.~Soltani.
\newblock Look, focus, act: Efficient and robust robot learning via human gaze
  and foveated vision transformers.
\newblock In \emph{IEEE International Conference on Robotics and Automation
  (ICRA)}, 2026.

\bibitem[Dass et~al.(2025)Dass, Hu, Abbatematteo, Stone, and
  Mart\'in-Mart\'in]{learningtolook2024}
S.~Dass, J.~Hu, B.~Abbatematteo, P.~Stone, and R.~Mart\'in-Mart\'in.
\newblock Learning to look: Seeking information for decision making via policy
  factorization.
\newblock In \emph{Conference on Robot Learning (CoRL)}, 2024.

\bibitem[{Dexmal}(2026)]{dm05}
{Dexmal}.
\newblock {DM0.5}.
\newblock \href{https://www.dexmal.com/blog/dm0.5?lang=en-US}{Technical blog},
  2026.

\bibitem[Gervet et~al.(2023)Gervet, Xian, Gkanatsios, and
  Fragkiadaki]{act3d2023}
T.~Gervet, Z.~Xian, N.~Gkanatsios, and K.~Fragkiadaki.
\newblock {Act3D}: {3D} feature field transformers for multi-task robotic
  manipulation.
\newblock In \emph{Conference on Robot Learning (CoRL)}, 2023.

\bibitem[Goyal et~al.(2024)Goyal, Blukis, Xu, Guo, Chao, and Fox]{rvt22024}
A.~Goyal, V.~Blukis, J.~Xu, Y.~Guo, Y.-W. Chao, and D.~Fox.
\newblock {RVT-2}: Learning precise manipulation from few demonstrations.
\newblock In \emph{Robotics: Science and Systems (RSS)}, 2024.

\bibitem[Grimaldi et~al.(2026)Grimaldi, Klee, Chen, Jian, Lee, Lu, Yu, and
  Nabi]{touch2trace2026}
M.~Grimaldi, D.~Klee, Z.~Chen, T.~Jian, W.~Lee, W.~Lu, T.~Yu, and S.~Nabi.
\newblock {Touch2Trace}: Tactile-driven imitation learning for dexterous cable
  tracing.
\newblock In \emph{Conference on Robot Learning (CoRL)}, 2026.

\bibitem[Hsieh et~al.(2026)Hsieh, Hsieh, Wang, Lin, Malik, Sreenath, and
  Qi]{dexscrew2026}
E.~Hsieh, W.-H. Hsieh, Y.-J. Wang, T.~Lin, J.~Malik, K.~Sreenath, and H.~Qi.
\newblock Learning dexterous manipulation skills from imperfect simulations.
\newblock In \emph{IEEE International Conference on Robotics and Automation
  (ICRA)}, 2026.

\bibitem[Huang et~al.(2025)Huang, Xu, Akinola, Yang, Sundaralingam, O'Flaherty,
  Fox, Wang, Mousavian, Chao, and Li]{vtrefine2025}
B.~Huang, J.~Xu, I.~Akinola, W.~Yang, B.~Sundaralingam, R.~O'Flaherty, D.~Fox,
  X.~Wang, A.~Mousavian, Y.-W. Chao, and Y.~Li.
\newblock {VT-Refine}: Learning bimanual assembly with visuo-tactile feedback
  via simulation fine-tuning.
\newblock In \emph{Conference on Robot Learning (CoRL)}, 2025.

\bibitem[James and Davison(2022)]{arm2022}
S.~James and A.~J. Davison.
\newblock {Q}-attention: Enabling efficient learning for vision-based robotic
  manipulation.
\newblock \emph{IEEE Robotics and Automation Letters (RA-L)}, 2022.

\bibitem[James et~al.(2022)James, Wada, Laidlow, and Davison]{c2fq2022}
S.~James, K.~Wada, T.~Laidlow, and A.~J. Davison.
\newblock Coarse-to-fine {Q}-attention: Efficient learning for visual robotic
  manipulation via discretisation.
\newblock In \emph{IEEE/CVF Conference on Computer Vision and Pattern
  Recognition (CVPR)}, 2022.

\bibitem[Kerr et~al.(2025)Kerr, Hari, Weber, Kim, Yi, bonnen, Goldberg, and
  Kanazawa]{eyerobot2025}
J.~Kerr, K.~Hari, E.~Weber, C.~M. Kim, B.~Yi, t.~bonnen, K.~Goldberg, and
  A.~Kanazawa.
\newblock Eye, robot: Learning to look to act with a {BC-RL} perception-action
  loop.
\newblock In \emph{Conference on Robot Learning (CoRL)}, 2025.

\bibitem[Kim et~al.(2021)Kim, Ohmura, and Kuniyoshi]{gaze2021}
H.~Kim, Y.~Ohmura, and Y.~Kuniyoshi.
\newblock Gaze-based dual resolution deep imitation learning for high-precision
  dexterous robot manipulation.
\newblock \emph{IEEE Robotics and Automation Letters (RA-L)}, 2021.

\bibitem[Li et~al.(2025{\natexlab{a}})Li, Li, Liu, Li, and
  Huang]{maniptrans2025}
K.~Li, P.~Li, T.~Liu, Y.~Li, and S.~Huang.
\newblock {ManipTrans}: Efficient dexterous bimanual manipulation transfer via
  residual learning.
\newblock In \emph{IEEE/CVF Conference on Computer Vision and Pattern
  Recognition (CVPR)}, 2025.

\bibitem[Li et~al.(2026)Li, Zhang, Luo, Yang, Wang, Zhang, Yu, Gao, Xue, Zhou,
  Zhu, Ding, Shen, and Xu]{lingbot}
L.~Li, Q.~Zhang, Y.~Luo, S.~Yang, R.~Wang, L.~Zhang, M.~Yu, Z.~Gao, N.~Xue,
  B.~Zhou, X.~Zhu, M.~Ding, Y.~Shen, and Y.~Xu.
\newblock Causal world modeling for robot control.
\newblock In \emph{Robotics: Science and Systems (RSS)}, 2026.

\bibitem[Li et~al.(2025{\natexlab{b}})Li, Wu, Xi, Li, Huang, Zhang, Chen, Wang,
  Zhu, Liu, and Huang]{controlvla2025}
P.~Li, Y.~Wu, Z.~Xi, W.~Li, Y.~Huang, Z.~Zhang, Y.~Chen, J.~Wang, S.-C. Zhu,
  T.~Liu, and S.~Huang.
\newblock {ControlVLA}: Few-shot object-centric adaptation for pre-trained
  vision-language-action models.
\newblock In \emph{Conference on Robot Learning (CoRL)}, 2025.

\bibitem[Liang et~al.(2024)Liang, Thomason, and B{\i}y{\i}k]{visarl2024}
A.~Liang, J.~Thomason, and E.~B{\i}y{\i}k.
\newblock {ViSaRL}: Visual reinforcement learning guided by human saliency.
\newblock In \emph{IEEE/RSJ International Conference on Intelligent Robots and
  Systems (IROS)}, 2024.

\bibitem[Lin et~al.(2025)Lin, Zhang, Li, Qi, Yi, Levine, and Malik]{hato2025}
T.~Lin, Y.~Zhang, Q.~Li, H.~Qi, B.~Yi, S.~Levine, and J.~Malik.
\newblock Learning visuotactile skills with two multifingered hands.
\newblock In \emph{IEEE International Conference on Robotics and Automation
  (ICRA)}, 2025.

\bibitem[Lyu et~al.(2025)Lyu, Chen, Du, Zhu, Liu, Wang, and
  Wang]{scissorbot2024}
J.~Lyu, Y.~Chen, T.~Du, F.~Zhu, H.~Liu, Y.~Wang, and H.~Wang.
\newblock {ScissorBot}: Learning generalizable scissor skill for paper cutting
  via simulation, imitation, and {Sim2Real}.
\newblock In \emph{Conference on Robot Learning (CoRL)}, 2024.

\bibitem[Namiki and Yokosawa(2021)]{origami2021}
A.~Namiki and S.~Yokosawa.
\newblock Origami folding by multifingered hands with motion primitives.
\newblock \emph{Cyborg and Bionic Systems}, 2021.

\bibitem[Peebles and Xie(2023)]{dit2023}
W.~Peebles and S.~Xie.
\newblock Scalable diffusion models with transformers.
\newblock In \emph{IEEE/CVF International Conference on Computer Vision
  (ICCV)}, 2023.

\bibitem[Pei et~al.(2026)Pei, Wu, Zhang, Luo, and Song]{inhandassembly2026}
L.~Pei, T.~Wu, H.~Zhang, P.~Luo, and J.~Song.
\newblock Assembling two parts in one hand.
\newblock In \emph{Conference on Robot Learning (CoRL)}, 2026.

\bibitem[Qi et~al.(2023)Qi, Yi, Suresh, Lambeta, Ma, Calandra, and
  Malik]{rotateit2023}
H.~Qi, B.~Yi, S.~Suresh, M.~Lambeta, Y.~Ma, R.~Calandra, and J.~Malik.
\newblock General in-hand object rotation with vision and touch.
\newblock In \emph{Conference on Robot Learning (CoRL)}, 2023.

\bibitem[Saxena et~al.(2023)Saxena, Sharma, and Kroemer]{mrest2023}
S.~Saxena, M.~Sharma, and O.~Kroemer.
\newblock Multi-resolution sensing for real-time control with vision-language
  models.
\newblock In \emph{Conference on Robot Learning (CoRL)}, 2023.

\bibitem[Shaw et~al.(2025)Shaw, Li, Yang, Srirama, Liu, Xiong, Mendonca, and
  Pathak]{bidex2025}
K.~Shaw, Y.~Li, J.~Yang, M.~K. Srirama, R.~Liu, H.~Xiong, R.~Mendonca, and
  D.~Pathak.
\newblock Bimanual dexterity for complex tasks.
\newblock In \emph{Conference on Robot Learning (CoRL)}, 2024.

\bibitem[Suresh et~al.(2024)Suresh, Qi, Wu, Fan, Pineda, Lambeta, Malik,
  Kalakrishnan, Calandra, Kaess, Ortiz, and Mukadam]{neuralfeels2024}
S.~Suresh, H.~Qi, T.~Wu, T.~Fan, L.~Pineda, M.~Lambeta, J.~Malik,
  M.~Kalakrishnan, R.~Calandra, M.~Kaess, J.~Ortiz, and M.~Mukadam.
\newblock {NeuralFeels} with neural fields: Visuotactile perception for in-hand
  manipulation.
\newblock \emph{Science Robotics}, 2024.

\bibitem[Tang et~al.(2026)Tang, Ji, Xing, Hao, Xu, Shao, Lu, Yu, Pang, and
  Zhang]{tang2026humanlike}
T.~Tang, X.~Ji, W.~Xing, C.~Hao, W.~Xu, L.~Shao, C.~Lu, Q.~Yu, J.~Pang, and
  K.~Zhang.
\newblock Towards human-like manipulation through {RL}-augmented teleoperation
  and mixture-of-dexterous-experts {VLA}.
\newblock In \emph{IEEE/RSJ International Conference on Intelligent Robots and
  Systems (IROS)}, 2026.

\bibitem[Wang et~al.(2024)Wang, Shi, Wang, Zhang, Fei-Fei, and Liu]{dexcap2024}
C.~Wang, H.~Shi, W.~Wang, R.~Zhang, L.~Fei-Fei, and K.~Liu.
\newblock {DexCap}: Scalable and portable mocap data collection system for
  dexterous manipulation.
\newblock In \emph{Robotics: Science and Systems (RSS)}, 2024.

\bibitem[Wang et~al.(2026{\natexlab{a}})Wang, Zhuang, Kragic, and
  Pokorny]{palm2026}
R.~Wang, Z.~Zhuang, D.~Kragic, and F.~T. Pokorny.
\newblock {PALM}: Enhanced generalizability for local visuomotor policies via
  perception alignment.
\newblock \emph{IEEE Robotics and Automation Letters (RA-L)}, 2026.

\bibitem[Wang et~al.(2026{\natexlab{b}})Wang, Huang, Li, Zhang, Liang, Jin,
  Chen, Chi, Zhou, Yu, Wang, Rui, Yao, Yuan, Shen, Zhu, Zhu, Gao, Chi, He,
  Zhang, Dong, Shao, and Zhao]{wang2026openwam}
Y.~Wang, S.~Huang, M.~Li, C.~Zhang, J.~Liang, W.~Jin, Y.~Chen, X.~Chi, D.~Zhou,
  Q.~Yu, Y.-K. Wang, Y.~Rui, S.~Yao, Z.~Yuan, Z.~Shen, K.~Zhu, Z.~Zhu, N.~Gao,
  X.~Chi, G.~He, S.~Zhang, H.~Dong, L.~Shao, and H.~Zhao.
\newblock {OpenWAM}: An open, modular exploration towards systematic
  world-action model pretraining.
\newblock In \emph{Conference on Robot Learning (CoRL), TAG Workshop}, 2026.

\bibitem[Xu et~al.(2025)Xu, Zhang, Hou, Xu, Fan, Veloso, and Song]{dexumi2025}
M.~Xu, H.~Zhang, Y.~Hou, Z.~Xu, L.~Fan, M.~Veloso, and S.~Song.
\newblock {DexUMI}: Using human hand as the universal manipulation interface
  for dexterous manipulation.
\newblock In \emph{Conference on Robot Learning (CoRL)}, 2025.

\bibitem[Xu et~al.(2026)Xu, Li, Wu, Qiu, and Shao]{fingereye2026}
Z.~Xu, Y.~Li, X.~Wu, T.~Qiu, and L.~Shao.
\newblock {FingerEye}: Learning dexterous manipulation with continuous
  vision-tactile sensing.
\newblock In \emph{Conference on Robot Learning (CoRL)}, 2026.

\bibitem[Yang et~al.(2025)Yang, Liu, Qin, Ding, Li, Cheng, Yang, Yi, and
  Wang]{ace2025}
S.~Yang, M.~Liu, Y.~Qin, R.~Ding, J.~Li, X.~Cheng, R.~Yang, S.~Yi, and X.~Wang.
\newblock {ACE}: A cross-platform and visual-exoskeletons system for low-cost
  dexterous teleoperation.
\newblock In \emph{Conference on Robot Learning (CoRL)}, 2024.

\bibitem[Yuan et~al.(2026)Yuan, Liu, Krishna, and Held]{grasp2dexterity2026}
Y.~Yuan, X.~Liu, S.~Krishna, and D.~Held.
\newblock From grasps to dexterity: Large-scale grasp pretraining for dexterous
  manipulation.
\newblock In \emph{Conference on Robot Learning (CoRL)}, 2026.

\bibitem[Zhang and Sennrich(2019)]{rmsnorm2019}
B.~Zhang and R.~Sennrich.
\newblock Root mean square layer normalization.
\newblock In \emph{Advances in Neural Information Processing Systems
  (NeurIPS)}, 2019.

\bibitem[Zhang et~al.(2026{\natexlab{a}})Zhang, Xu, Dhafer, Yue, Dong, and
  Hao]{iss2026}
H.~Zhang, M.~Xu, A.~Dhafer, S.~Yue, H.~Dong, and Z.~D. Hao.
\newblock Embodied interpretability: Linking causal understanding to
  generalization in vision-language-action models.
\newblock In \emph{International Conference on Machine Learning (ICML)}, 2026.

\bibitem[Zhang et~al.(2026{\natexlab{b}})Zhang, Wu, Chen, Wang, Zhong, Zhang,
  Yang, and Chen]{dexknot2026}
J.~Zhang, R.~Wu, H.~Chen, Y.~Wang, Y.~Zhong, C.~Zhang, Y.~Yang, and Y.~Chen.
\newblock {DexKnot}: Generalizable visuomotor policy learning for dexterous
  bag-knotting manipulation.
\newblock In \emph{IEEE International Conference on Robotics and Automation
  (ICRA)}, 2026.

\bibitem[Zhu et~al.(2023{\natexlab{a}})Zhu, Jiang, Stone, and Zhu]{groot2023}
Y.~Zhu, Z.~Jiang, P.~Stone, and Y.~Zhu.
\newblock Learning generalizable manipulation policies with object-centric {3D}
  representations.
\newblock In \emph{Conference on Robot Learning (CoRL)}, 2023.

\bibitem[Zhu et~al.(2023{\natexlab{b}})Zhu, Joshi, Stone, and Zhu]{viola2023}
Y.~Zhu, A.~Joshi, P.~Stone, and Y.~Zhu.
\newblock {VIOLA}: Imitation learning for vision-based manipulation with object
  proposal priors.
\newblock In \emph{Conference on Robot Learning (CoRL)}, 2022.

\bibitem[Zhuang et~al.(2026)Zhuang, Wang, Heppert, Hahn, Valada, Pokorny, and
  Kragic]{seeker2026}
Z.~Zhuang, R.~Wang, N.~Heppert, J.~F. Hahn, A.~Valada, F.~T. Pokorny, and
  D.~Kragic.
\newblock Attention from action, for action: Emergent visual bottlenecks for
  policy learning.
\newblock In \emph{Conference on Robot Learning (CoRL)}, 2026.

\end{thebibliography}
} 

\clearpage
\maketitlesupplementary
\appendix
\section{Method Details}
\label{sec:method_details}

\subsection{Offline AVR Discovery}
\label{sec:roi_example_settings}

\paragraph{From $d_j$ to the sensitivity map.}
The normalized action chunks $A^{(0)}$ and $A^{(j)}$ are generated using identical initial noise and ten flow-matching denoising steps. Following \cref{eq:sensitivity}, the action discrepancy $d_j$ is computed over the first $K=20$ horizon steps across all $|\mathcal I_{\mathrm{act}}|=65$ active DoFs. Given a $16\times16$ grid of perturbation locations, we accumulate each score weighted by its soft spatial mask $M_j$, normalize by total mask coverage to account for window overlap, and apply Gaussian smoothing $G_\sigma$, where $\sigma$ is measured in pixels on the discovery image:
\begin{equation}
\begin{gathered}
 \widetilde{\mathcal S}_t(u,v)=
 \frac{\sum_{j=1}^{256}d_j M_j(u,v)}{\max\!\left(\sum_{j=1}^{256}M_j(u,v),\,10^{-6}\right)},\\
 F_t=G_{10}(\widetilde{\mathcal S}_t),\\
 \mathcal S_t=\frac{F_t-\min F_t}{\max F_t-\min F_t}.
\end{gathered}
 \label{eq:sensitivity_map}
\end{equation}
Min--max normalization maps valid responses to $\mathcal S_t\in[0,1]$; when $\max F_t-\min F_t\leq 10^{-9}$, we set $\mathcal S_t=0$.

\paragraph{From the sensitivity map to the crop.}
Let $T_t=G_2(\mathcal S_t)$ denote a lightly smoothed version of $\mathcal S_t$. We extract the 8-connected high-response region containing the global peak:
\begingroup
\footnotesize
\begin{equation}
\begin{gathered}
 \theta_t=\max\!\left(Q_{0.85}\!\left(\{T_t(u,v):T_t(u,v)>0\}\right),\,0.8\max T_t\right),\\
 \mathcal C_t=\operatorname{CC}_8\!\left(\{(u,v):T_t(u,v)\geq\theta_t\},\arg\max T_t\right),
\end{gathered}
 \label{eq:avr_peak_component}
\end{equation}
\endgroup
where $Q_{0.85}$ denotes the 85th percentile over strictly positive entries of $T_t$. Letting $w_{\mathcal C_t}$ and $h_{\mathcal C_t}$ be the pixel width and height of $\mathcal C_t$'s bounding box on the dense sensitivity map, we compute the response-weighted center \mbox{and padded extents}:
\begin{equation}
\begin{gathered}
 (\bar u_t,\bar v_t)=\frac{\sum_{(u,v)\in\mathcal C_t}T_t(u,v)^2(u,v)}{\sum_{(u,v)\in\mathcal C_t}T_t(u,v)^2},\\
 (w,h)=\operatorname{clip}\!\left((w_{\mathcal C_t},h_{\mathcal C_t})+24,\,88,\,156\right).
\end{gathered}
 \label{eq:avr_crop_extents}
\end{equation}
This adds a 12-pixel margin on each side and clips width and height independently. We center the box at $(\bar u_t,\bar v_t)$ and shift it within the discovery image. The box is first mapped to native head-image coordinates and rounded to integers. In these native-image coordinates, it is then made square using its longer side, expanded by $1.2$, and shifted within image boundaries. Let $W_I$ and $H_I$ be the native image width and height. The resulting pixel-edge box $B_t=(x_0,y_0,x_1,y_1)$ defines the normalized Teacher AVR and the local visual crop used for policy training:
\begin{equation}
\begin{gathered}
 b_t=\left(\frac{x_0}{W_I},\,\frac{y_0}{H_I},\,\frac{x_1}{W_I},\,\frac{y_1}{H_I}\right),\\
 C_t=\operatorname{Resize}^{\mathrm{bicubic}}_{224\times224}
 \!\left(I_t^{\mathrm{head}}[y_0:y_1,x_0:x_1]\right),
\end{gathered}
 \label{eq:avr_teacher_crop}
\end{equation}

\subsection{AVR Localization Module}

\paragraph{Architecture and parameterization.}
Let $\{h_{t,i}^{\mathrm{pre}}\}_{i=1}^{N}$ denote the contextualized tokens of the frozen VLM global prefix, comprising the global views, language, and state (excluding local-crop tokens). Applying LayerNorm, learned attention pooling, and a 256-unit GELU MLP yields the predicted box $\hat b_t$ in normalized head-image coordinates through the following constrained parameterization:
\begin{equation}
\begin{gathered}
 \widetilde h_{t,i}=\operatorname{LN}(h_{t,i}^{\mathrm{pre}}),\\
 \alpha_{t,i}=\frac{\exp(w_p^{\mathsf T}\widetilde h_{t,i})}{\sum_{j=1}^{N}\exp(w_p^{\mathsf T}\widetilde h_{t,j})},\\
 q_t=W_2\operatorname{GELU}\!\left(W_1\sum_{i=1}^{N}\alpha_{t,i}\widetilde h_{t,i}+a_1\right)+a_2,\\
 e_t=0.01+0.98\sigma(q_{t,3:4}),\\
 \mu_t=\frac{e_t}{2}+(1-e_t)\odot\sigma(q_{t,1:2}),\\
 \hat b_t=\left[\mu_t-\frac{e_t}{2};\,\mu_t+\frac{e_t}{2}\right].
\end{gathered}
 \label{eq:avr_localization_model}
\end{equation}
Here $\sigma$ is the sigmoid function, $e_t$ and $\mu_t$ represent the predicted box extent and center, and $\hat b_t$ stores normalized corner coordinates. Parameterizing $\mu_t$ and $e_t$ from $q_{t,1:2}$ and $q_{t,3:4}$ guarantees that $\hat b_t$ remains valid within $[0,1]^4$.

\paragraph{Training objective.}
Given the Teacher AVR $b_t$ from \cref{sec:teacher}, we train the localization module over valid samples $\mathcal V$ using a sum of Smooth-L1 coordinate loss \mbox{and GIoU loss:}
\begingroup
\small
\begin{gather}
 \mathcal L_{\mathrm{loc}}=
 \frac{1}{\max(1,|\mathcal V|)}\sum_{t\in\mathcal V}\Biggl[\frac{1}{4}\sum_{i=1}^{4}\operatorname{SmoothL1}_{0.1}(\hat b_{t,i}-b_{t,i})\notag\\
 {}+1-\operatorname{GIoU}(\hat b_t,b_t)\Biggr].
 \label{eq:localization_loss}
\end{gather}
\endgroup
Here $\operatorname{SmoothL1}_{0.1}$ uses transition threshold $0.1$ and is averaged over the four coordinates, while the GIoU term has unit weight. During localization training, the VLM backbone, action expert, and \mbox{local conditioning branch remain frozen.}

\subsection{Action-conditioned Visual Modulation}

\paragraph{Local visual feature adapter.}
The shared SigLIP encoder $E_{\mathrm{SigLIP}}$ encodes $C_t$. An antialiased bilinear resize $R_{4\times4}$ downsamples the resulting spatial token grid to $4\times4$ tokens, followed by a linear projection and mean pooling:
\begin{equation}
\begin{gathered}
 v_{t,k}=\left[R_{4\times4}\!\left(E_{\mathrm{SigLIP}}(C_t)\right)\right]_k,\\
 z_{t,k}=W_v v_{t,k}+a_v,\\
 \bar z_t=\frac{1}{16}\sum_{k=1}^{16}z_{t,k},
\end{gathered}
 \label{eq:local_visual_adapter}
\end{equation}
where $z_{t,k}\in\mathbb R^{d_c}$ and $d_c=1,024$ matches the action-expert and flow-time embedding dimension.

\paragraph{Geometry embedding.}
From the normalized AVR corners $b_t=(x_0,y_0,x_1,y_1)$ and crop-validity flag $m_t$, we extract the center and size and project them into $\mathbb R^{d_c}$:
\begin{equation}
\begin{gathered}
 c_x=\frac{x_0+x_1}{2},\quad c_y=\frac{y_0+y_1}{2},\\
 w=x_1-x_0,\quad h=y_1-y_0,\\
 g_t=W_b[c_x,c_y,w,h,m_t]^{\mathsf T}+a_b\in\mathbb R^{d_c}.
\end{gathered}
 \label{eq:geometry_embedding}
\end{equation}

\paragraph{AVR condition projector.}
A two-layer MLP fuses appearance and geometry while bounding the output \mbox{RMS by $1$:}
\begin{equation}
\begin{gathered}
 r_t=W_{p,2}\operatorname{SiLU}\!\left(W_{p,1}[\bar z_t;g_t]+a_{p,1}\right)+a_{p,2},\\
 f_{\mathrm{AVR}}(\bar z_t,b_t)=\frac{r_t}{\max\!\left(1,\,\sqrt{\frac{1}{d_c}\sum_{i=1}^{d_c}r_{t,i}^{2}}\right)}.
\end{gathered}
 \label{eq:avr_magnitude_cap}
\end{equation}
This conditioning vector enters $c_{t,\tau}$ via the learnable gate $\tanh(\gamma)$ in \cref{eq:condition} (initialized at $\gamma=0$). When a crop is invalid or randomly dropped during training (with probability $0.3$), we substitute a learned null representation prior to RMS capping. At inference time, $(\hat C_t,\hat b_t)$ replaces $(C_t,b_t)$, and the geometry features are computed from the coordinates used to crop $\hat C_t$.
\section{Experimental Setup}
\label{sec:experimental_details}

\subsection{Dataset Splits and Success Criteria}
\label{sec:dataset_splits_success}
We train task-specific policies on 510 paper-folding, 780 paper-cutting, 1,061 paper-unfolding, and 554 thread-cutting episodes. Task success criteria are defined as follows:
\begin{itemize}
    \item \textbf{Paper Folding:} The paper has been folded twice, producing two creases; the creases need not coincide with the marked lines during either of the \mbox{two folding operations.}
    \item \textbf{Paper Cutting:} The length of the target black contour cut within a $1\,\mathrm{mm}$ deviation must be at least 70\% of its total length, for both straight and curved contours. The designated piece must also be completely separated from the remaining paper sheet and not merely partially cut along its boundary.
    \item \textbf{Paper Unfolding:} The paper sheet is fully unfolded at the end of the rollout.
    \item \textbf{Thread Cutting:} The thread across the stand is completely severed into two pieces.
\end{itemize}

\subsection{Training Schedules and Baseline Implementations}
\label{sec:training_baselines}

\paragraph{Optimization schedules.}
The reference policy, action policy, and localization module use FSDP and AdamW ($(\beta_1,\beta_2)=(0.9,0.95)$, $\epsilon=10^{-8}$, weight decay $10^{-10}$, gradient clip $1.0$) with 1,000-step linear warmup to $2.5\times10^{-5}$ and cosine decay to $2.5\times10^{-6}$. The data, hardware, batch sizes, and schedules are specified below for reference-policy optimization, action learning, and subsequent localization-module fitting:
\begin{itemize}
    \item \textbf{Reference policy:} Initialized from $\pi_{0.5}$~\cite{pi05} and trained without local crops on the same demonstration sets as the action policy (\cref{sec:dataset_splits_success}; 16 PPU-ZW810E, batch size 16), running for 2 epochs per task.
    \item \textbf{Action policy:} Initialized from $\pi_{0.5}$ and trained on the full datasets (\cref{sec:dataset_splits_success}) with offline teacher crops, jointly updating the VLM backbone, shared SigLIP encoder, local conditioning modules, and action expert. Uses batch size 16 across 16 PPU-ZW810E, running 2 epochs per task. Crop dropout is $0.3$; spatial jittering (probability $0.5$) applies translations up to $\pm 5\%$ of box size, scale in $[0.9,1.1]$, and rotations up to $\pm 5^\circ$.
    \item \textbf{Localization module:} Trained on the same episodes in~\cref{sec:dataset_splits_success} using Teacher AVR supervision. Starting from the converged action policy, only the localization head is updated with all policy weights frozen for 8 epochs (16 PPU-ZW810E, batch size 64, EMA disabled) under equally weighted Smooth-L1 (transition $0.1$) and GIoU losses.
\end{itemize}

\paragraph{Compared policies.}
LingBot-VA~\cite{lingbot}, OpenWAM~\cite{wang2026openwam}, DM0.5~\cite{dm05}, $\pi_0$~\cite{pi0}, and $\pi_{0.5}$~\cite{pi05} are trained separately per task on the identical demonstration sets (\cref{sec:dataset_splits_success}) using 16 PPUs, batch size 16, and 2 epochs.

\subsection{Ablation Architectures and Region Priors}
\label{sec:ablation_architectures}

\begin{figure*}[p]
  \centering
  \includegraphics[width=\textwidth]{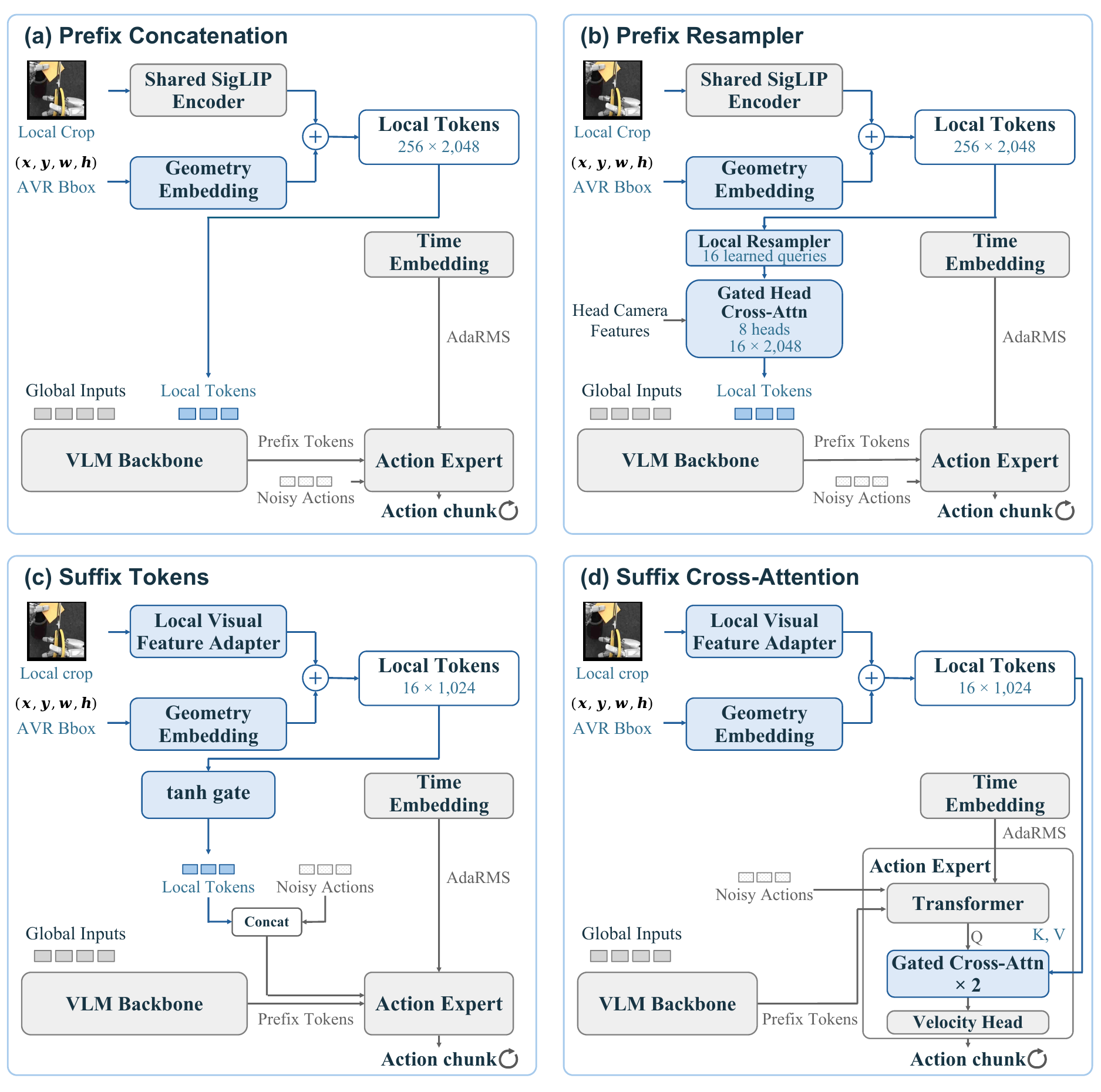}
  \caption{\textbf{Conditioning architectures.} Four alternatives to suffix AdaRMS in \cref{figs:pipeline}(c). The circled plus adds region geometry to local visual tokens. (a) Prefix Concatenation inserts 256 geometry-enriched tokens into the VLM prefix. (b) Prefix Resampler uses local resampling and gated head-camera cross-attention to produce 16 prefix tokens. (c) Suffix Tokens adds geometry after local feature adaptation, then gates the 16 tokens and places them before the noisy action tokens. (d) Suffix Cross-Attention applies two residual-gated attention layers after the final action Transformer block and before velocity prediction, using action states as queries ($Q$) and local tokens as keys and values ($K,V$). Local feature adapters include shared SigLIP encoding. Blue marks the local-conditioning path; gray marks existing policy components. Action chunks result from repeated flow-matching updates.}
  \label{fig:conditioning_architectures}
\end{figure*}

\paragraph{Conditioning architectures.}
The five AVR-based conditioning routes in \cref{tab:ablations} share the same datasets, trainable SigLIP encoder, and $224\times224$ crop inputs. The object-centric AdaRMS baseline is a separate region-supervision comparison, with the same action-conditioning architecture and a different source of region supervision. The four alternative conditioning routes are illustrated in \cref{fig:conditioning_architectures}.
\begin{itemize}
    \item \textbf{Prefix Concatenation.} Each crop is encoded into 256 tokens of width 2,048 and summed with a linear embedding of its four normalized corner coordinates. We insert these geometry-enriched tokens alongside the global image tokens in the VLM prefix so that bidirectional prefix attention can update the global representations before action generation. Any invalid or dropped crops are zeroed out and masked from attention, excluding their local features from the contextualized prefix representations used by the action expert to generate the next action chunk.
    \item \textbf{Prefix Resampler.} To avoid a long 256-token prefix, we first add the normalized bounding-box geometry embedding to the crop features, project them to width 512, and compress them into 16 tokens via 8-head cross-attention using 16 learned queries, followed by a residual GELU FFN (width 2,048). Next, a second cross-attention layer queries the full head-camera features to enrich this local summary, modulated by a zero-initialized $\tanh$ gate. Finally, the resulting 16 tokens are projected to width 2,048 and inserted into the VLM prefix, with invalid or dropped crops masked out.
    \item \textbf{Suffix Tokens.} We first downsample the crop features via antialiased bilinear resize into a $4\times4$ grid (16 tokens) and project them to dimension 1,024. Each token is then summed with a linear embedding of the normalized bounding-box geometry (center, width, height, and validity) and scaled by a zero-initialized $\tanh$ gate. Within the action suffix, the 16 local crop tokens precede the noisy action tokens, giving the order global prefix--local crop tokens--noisy action tokens: local tokens attend to the global prefix and to one another, while action tokens can additionally attend to the local tokens (the global prefix never attends back to the suffix). When a crop is invalid or dropped, we substitute 16 learned null tokens.
    \item \textbf{Suffix Cross-Attention.} We apply two consecutive gated cross-attention layers (width 1,024, 8 heads) after the final action transformer block and before the velocity projection head. Using the action hidden states as queries and the 16 geometry-enriched local tokens as keys and values, each layer applies separate normalization to the queries and context, computes scaled dot-product attention, and injects the output via a zero-initialized $\tanh$ residual gate. This refines action features without altering the transformer sequence length. Invalid or dropped crops use learned null context tokens as the keys and values in both attention layers for the current observation throughout action generation.
    \item \textbf{Suffix AdaRMS (Ours).} This route uses the local feature adapter, geometry embedding, and AVR condition projector defined in \cref{sec:adarms}. The object-centric variant retains these modules and replaces teacher AVRs with SAM3.1-derived boxes for both policy conditioning and localization supervision.
\end{itemize}

\paragraph{Shared training and deployment protocol.}
In all variants, each localization module is trained on top of its route's frozen action backbone using a \emph{crop-free global prefix}, with both teacher and predicted crop tokens excluded from the input. During policy training, crop dropout with probability $0.3$ zeros both the box coordinates and the validity flag, preventing geometry from revealing the position or size of a dropped region. At deployment, each route predicts a box from its crop-free prefix, then extracts and encodes the crop for prefix or suffix conditioning. 

\paragraph{SAM3.1 region selection.}
SAM3.1 masks are converted to bounding boxes for the object-centric baseline. Thread cutting uses scissors masks, while paper folding and unfolding use paper masks. In paper cutting, segmentation targets the scissors initially and the paper later, providing task-specific region supervision for both policy conditioning and localization-module training using the same selected object-centric region targets.

\subsection{Cross-Table Evaluation Consistency}
\label{sec:cross_table_consistency}
\cref{tab:baseline} and \cref{tab:ablations} evaluate the same Focus4Dex checkpoint across two separate collection campaigns conducted at different times, with minor variations in lighting and tool placement. Within each table, all methods on a given task were evaluated within a shared time window under identical settings; accordingly, success rates should be compared within each table rather than across tables.

\section{Failure Analysis}
\label{sec:additional_results}

\cref{fig:qualitative_all_models} shows representative successful and failed real-world rollouts of our method and the baselines on all four tasks. Each row contains eight chronological head-camera frames, and an inset in the seventh frame marks the final outcome. All four tasks share three core requirements: establishing the intended contact (\eg, grasping the paper or scissors), maintaining the necessary support or grasp throughout the motion, and reaching the required geometry and final state (\eg, aligning the blades, tracking the contour, and completing separation). We therefore group failures into three categories: (1) \emph{failure to establish contact}; (2) \emph{failure to maintain contact}, \eg, paper slippage, weak support from the coordinated hand, or loss of the scissor grasp; and (3) \emph{failure to reach the goal geometry}, \eg, misalignment, contour deviation, or incomplete separation. As the folding and unfolding examples show, binary success metrics do not fully capture execution quality. In \cref{fig:qualitative_all_models}, the first row of each task is a successful rollout, and the remaining rows follow this category order.

\begingroup
\setlength{\parindent}{0pt}
\newcommand{\rolloutimagewidth}{\textwidth}
\newcommand{\rolloutrowgap}{0.5pt}

\newcommand{\rolloutsingle}[2]{%
  \begin{minipage}{\linewidth}%
    \raisebox{-0.5\height}{\makebox[0pt][r]{\scriptsize #1\hspace{8pt}}}%
    \raisebox{-0.5\height}{\includegraphics[page=#2,width=\rolloutimagewidth]{figs/rollout_baselines_cleanfinal2_20261124.pdf}}%
  \end{minipage}\par\vspace{\rolloutrowgap}}

\newcommand{\rolloutadditional}[2]{%
  \begin{minipage}{\linewidth}%
    \raisebox{-0.5\height}{\makebox[0pt][r]{\scriptsize #1\hspace{8pt}}}%
    \raisebox{-0.5\height}{\includegraphics[page=#2,width=\rolloutimagewidth]{figs/rollout_additional_four_20261124.pdf}}%
  \end{minipage}\par\vspace{\rolloutrowgap}}
\newcommand{\rolloutcurve}[2]{%
  \begin{minipage}{\linewidth}%
    \raisebox{-0.5\height}{\makebox[0pt][r]{\scriptsize #1\hspace{8pt}}}%
    \raisebox{-0.5\height}{\includegraphics[page=#2,width=\rolloutimagewidth]{figs/rollout_curve_cleanstart_20261124.pdf}}%
  \end{minipage}\par\vspace{\rolloutrowgap}}
\newcommand{\rolloutstraightclean}[2]{%
  \begin{minipage}{\linewidth}%
    \raisebox{-0.5\height}{\makebox[0pt][r]{\scriptsize #1\hspace{8pt}}}%
    \raisebox{-0.5\height}{\includegraphics[page=#2,width=\rolloutimagewidth]{figs/rollout_straight_openwam_cleanstart_20261124.pdf}}%
  \end{minipage}\par\vspace{\rolloutrowgap}}
\newcommand{\rollouttask}[1]{\par\vspace{0.5pt}{\centering\footnotesize #1\par}\vspace{2.5pt}}
\begin{figure*}[p]
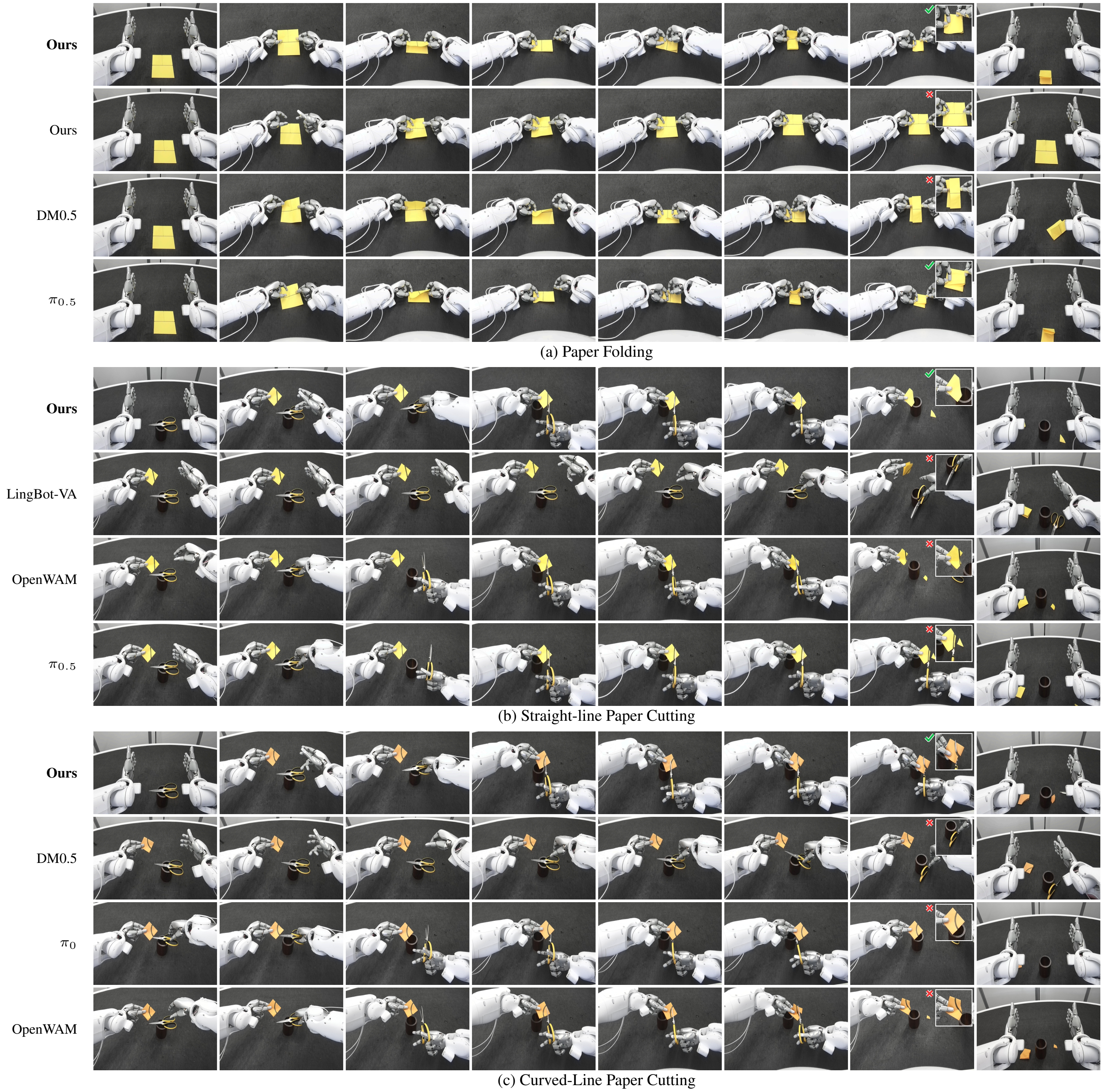

\centering
\rolloutsingle{\textbf{Ours}}{7}
\rolloutadditional{Ours}{4}
\rolloutsingle{DM0.5}{9}
\rolloutsingle{$\pi_{0.5}$}{8}
\rollouttask{(a) Paper Folding}
\rolloutsingle{\textbf{Ours}}{5}
\rolloutadditional{LingBot-VA}{2}
\rolloutstraightclean{OpenWAM}{1}
\rolloutadditional{$\pi_{0.5}$}{3}
\rollouttask{(b) Straight-line Paper Cutting}
\rolloutsingle{\textbf{Ours}}{6}
\rolloutadditional{DM0.5}{1}
\rolloutcurve{$\pi_0$}{1}
\rolloutcurve{OpenWAM}{2}
\rollouttask{(c) Curved-Line Paper Cutting}
\caption{\textbf{Qualitative Model Showcases (a--c).} Numbers in parentheses denote the failure categories defined in \cref{sec:additional_results}; the first row of each task is a successful rollout of our method. \textbf{(a) Paper Folding.} From top to bottom: Ours completes both folds; another Ours rollout fails to grasp the paper (1); DM0.5 lets the paper slip during pressing (2); and $\pi_{0.5}$ completes both folds, but the folded edges are not well aligned. \textbf{(b) Straight-Line Paper Cutting.} Ours cuts along the target line; LingBot-VA fails to pick up the scissors (1); and OpenWAM and $\pi_{0.5}$ deviate from the marked line (3). \textbf{(c) Curved-Line Paper Cutting.} Ours follows the target curve; DM0.5 fails to pick up the scissors (1); $\pi_0$ misaligns the blades with the paper (3); and OpenWAM deviates from the marked contour (3).}
\label{fig:qualitative_all_models}
\end{figure*}
\clearpage
\begin{figure*}[p]
\centering
\ContinuedFloat
\rolloutsingle{\textbf{Ours}}{10}
\rolloutsingle{DM0.5}{13}
\rolloutsingle{$\pi_{0.5}$}{11}
\rolloutsingle{$\pi_0$}{12}
\rollouttask{(d) Paper Unfolding}
\rolloutsingle{\textbf{Ours}}{1}
\rolloutsingle{LingBot-VA}{4}
\rolloutsingle{DM0.5}{3}
\rolloutsingle{$\pi_{0.5}$}{2}
\rollouttask{(e) Thread Cutting}
\caption{\textbf{Qualitative Model Showcases (d--e).} Rows follow the same layout and failure categories (\cref{sec:additional_results}) as in (a--c). \textbf{(d) Paper Unfolding.} From top to bottom: Ours fully unfolds the sheet; DM0.5 fails to engage and lift the folded layer with its right hand (1); $\pi_{0.5}$ fails to hold the paper in place with its left index finger (2); and $\pi_0$ fully unfolds the sheet, although its left index finger does not provide stable support. \textbf{(e) Thread Cutting.} Ours severs the thread; LingBot-VA fails to pick up the scissors (1); DM0.5 loses its grasp on the scissors during cutting (2); and $\pi_{0.5}$ closes the blades without completely severing the thread (3).}

\end{figure*}
\endgroup
\par

\end{document}